\documentclass[10pt,twocolumn,letterpaper]{article}

\usepackage[pagenumbers]{iccv} 

\definecolor{iccvblue}{rgb}{0.21,0.49,0.74}
\usepackage[pagebackref,breaklinks,colorlinks,allcolors=iccvblue]{hyperref}
\usepackage{multirow}
\usepackage{arydshln}
\usepackage{pifont}
\usepackage{caption}
\usepackage{listings}
\usepackage{ragged2e}
\usepackage{adjustbox}
\usepackage{placeins}
\usepackage{stfloats}

\def\paperID{7818} 
\def\confName{ICCV}
\def\confYear{2025}

\title{Enhancing Image Restoration Transformer via \\
Adaptive Translation Equivariance}

\author{JiaKui Hu$^{1,2,3}$, Zhengjian Yao$^{1,2,3}$, Lujia Jin$^{4}$, Hangzhou He$^{1,2,3}$, Yanye Lu$^{1,2,3}$\thanks{Corresponding author.}\\
$^1$Institute of Medical Technology, Peking University Health Science Center, Peking University \\
$^2$Biomedical Engineering Department, College of Future Technology, Peking University \\
$^3$National Biomedical Imaging Center, Peking University\\ 
$^4$China Mobile Research Institute \\
{\tt\small jkhu29@stu.pku.edu.cn, yanye.lu@pku.edu.cn}
}

\begin{document}
\maketitle
\begin{abstract}
Translation equivariance is a fundamental inductive bias in image restoration, ensuring that translated inputs produce translated outputs. Attention mechanisms in modern restoration transformers undermine this property, adversely impacting both training convergence and generalization. To alleviate this issue, we propose two key strategies for incorporating translation equivariance: slide indexing and component stacking. Slide indexing maintains operator responses at fixed positions, with sliding window attention being a notable example, while component stacking enables the arrangement of translation-equivariant operators in parallel or sequentially, thereby building complex architectures while preserving translation equivariance. However, these strategies still create a dilemma in model design between the high computational cost of self-attention and the fixed receptive field associated with sliding window attention. To address this, we develop an adaptive sliding indexing mechanism to efficiently select key-value pairs for each query, which are then concatenated in parallel with globally aggregated key-value pairs. The designed network, called the Translation Equivariance Adaptive Transformer (TEAFormer), is assessed across a variety of image restoration tasks. The results highlight its superiority in terms of effectiveness, training convergence, and generalization.
\end{abstract}    
\vspace{-5mm}
\section{Introduction}
\label{sec:intro}





Image restoration aims to reconstruct high-quality (HQ) images from their low-quality (LQ) counterparts, which are often affected by degradations resulting from the imaging system or environment. As a constituent of the imaging system, image restoration methods should satisfy the following fidelity property: the restoration results for the target region should remain equivariant under geometric transformations of the input. This fidelity property is termed translation equivariance, which is one of the inherent inductive biases in image restoration.

\vspace{-3mm}
\begin{figure}[ht]
\centering
\includegraphics[width=\linewidth]{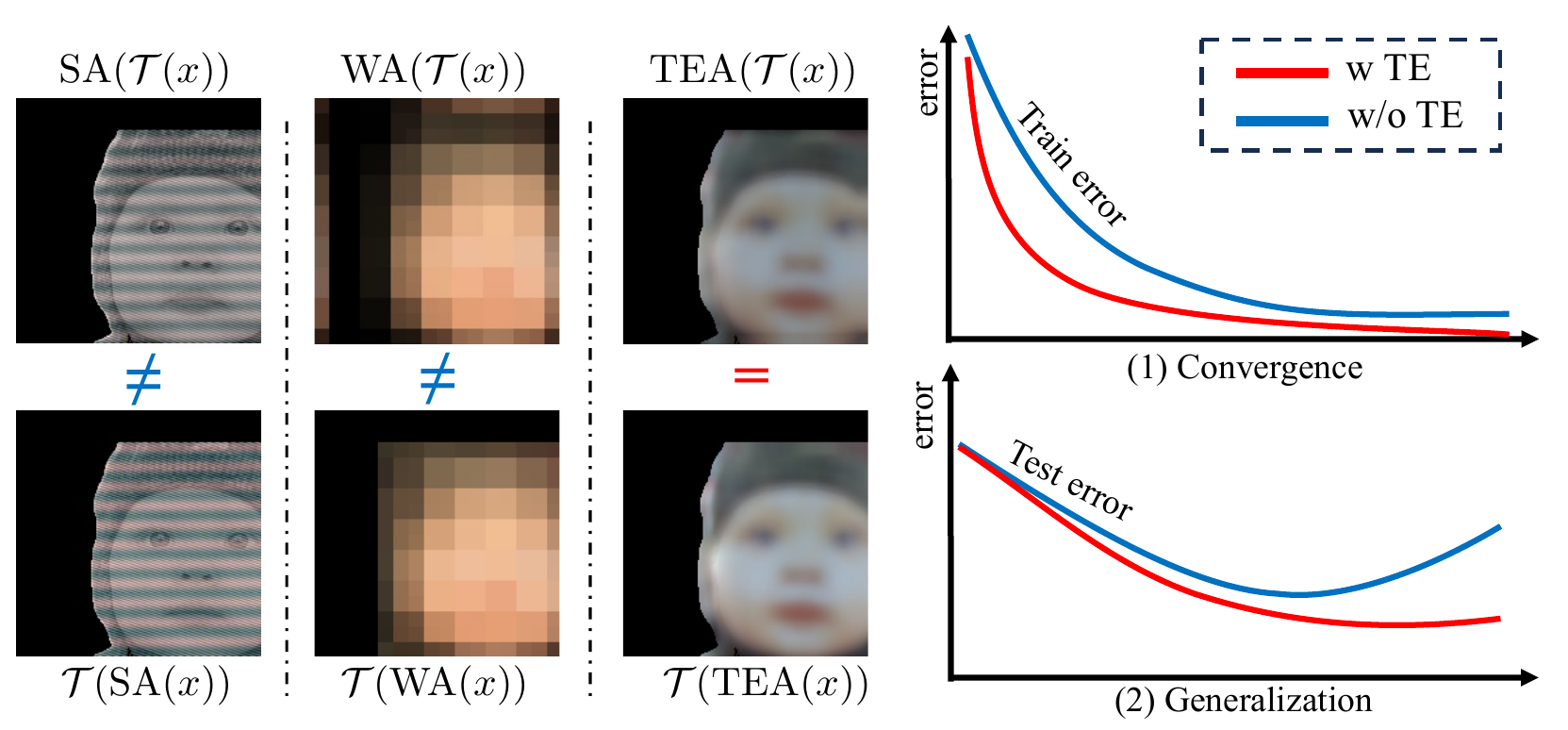}
\vspace{-8mm}
\caption{$\mathcal{T}(\cdot)$ means the translation function. "SA" and "WA" are commonly used for self-attention and window attention, respectively, but disrupt translation equivariance (TE) due to position encoding and feature shifting. "TEA" is our proposed translation equivariance adaptive attention, which satisfies TE. TE promotes faster convergence and better generalization.
}\label{fig:te_overview}
\vspace{-7mm}
\end{figure}

\begin{table}[ht]
\centering
\scalebox{0.7}{
\begin{tabular}{c|c|c|c|c|c}
\toprule[0.15em]
\multirow{2}{*}{\textbf{Attentions}} & \multirow{2}{*}{\textbf{TE}} & Performance & NTK~\cite{jacot2018neural} & SRGA~\cite{liu2023evaluating} & \multirow{2}{*}{Complexity} \\
\cline{3-5}
&  & PSNR$\uparrow$ / SSIM$\uparrow$ & Condition $\downarrow$ & Value $\downarrow$ & \\
\midrule[0.15em]
SA   & \ding{56} & 28.21 / 0.8435 & 11140.44 & 4.075 & $\mathcal{O}(N^2)$ \\
WA   & \ding{56} & 27.45 / 0.8254 & 1746.49 & 3.655 & $\mathcal{O}(N)$ \\
\specialrule{0em}{1pt}{1pt}
\hdashline
\specialrule{0em}{1pt}{1pt}
\textbf{TEA (Ours)} & \ding{52} & \textbf{28.67} / \textbf{0.8489} & \textbf{236.78} & \textbf{3.275} & $\mathcal{O}(N)$ \\
\bottomrule
\end{tabular}
}
\vspace{-3mm}
\caption{In $4\times$ image super-resolution, we study performance, NTK \cite{chen2021vision} convergence speed, SRGA \cite{liu2023evaluating} generalization, and computation complexities. We replace WA in SwinIR with SA and TEA for testing. TEA shows fast convergence, better generalization, and linear complexity. The results are tested on Urban100.}
\label{tab:te_ops}
\vspace{-3mm}
\end{table}

In restoration networks~\cite{lim2017enhanced,zhang2018rcan,zhang2021plug,Zamir_2021_CVPR_mprnet} based on convolution neural networks (CNNs), the paradigm of sliding information extraction inherent to convolutions~\cite{gordon2019convolutional} imparted a natural translation equivariance (TE)~\footnote{Translation equivariance and translation invariance are concepts that are frequently misconstrued. Our paper is dedicated to the analysis of equivariance. We have provided a clear delimitation and discussion concerning these concepts in the supplementary material.} property in restoration networks. However, as shown in Figure~\ref{fig:te_overview} and Table~\ref{tab:te_ops}, the advent of Self Attention (SA)~\cite{vaswani2017attention} and Window Attention (WA)~\cite{liu2021Swin} within Transformers has broken this fidelity property, leading to slow convergence and poor generalization in restoration transformers.



In this paper, we propose two fundamental strategies to bring TE back into the restoration transformers: \textbf{(1)} \textit{slide indexing} and \textbf{(2)} \textit{component stacking}. The slide indexing strategy employs a fixed container, \textit{e.g.}, a fixed rectangular window, to globally extract the input in a sliding fashion. This method requires the operators to calculate instantaneously, thereby maintaining the consistency of positional coordinates between the input and output of the model, which facilitates the integration of TE. The component stacking strategy involves stacking modules that satisfy the TE property in parallel or in series, thus enhancing architectures while ensuring TE in complex networks.

However, these foundational strategies present a persistent challenge in the design of restoration transformers. Selecting SA to achieve superior performance involves accepting its substantial computational complexity~\cite{Zamir2021Restormer}; alternatively, opting for sliding window attention provides linear complexity but results in a fixed receptive field~\cite{tian2024image} and reduced performance. To break this dilemma, we replace the global key-value pair indexing mechanism in vanilla SA with an adaptive slide indexing mechanism to design the Translation Equivariance Adaptive (TEA) attention. To retain the extraction of global information, TEA incorporates a downsampling self-attention branch without position encoding, which reduces the key-value pair's resolution by aggregating them, thereby providing TEA with coarse, yet effective global information.

In summary, our contributions are threefold. \textbf{First,} we investigate TE in image restoration, analyzing its impact on convergence and generalization. The results indicate that the integration of TE is essential in the design of restoration networks. \textbf{Second,} we propose TEA, which uses adaptive slide indexing methods to incorporate the TE property while preserving linear complexity and ensuring a global receptive field, thereby enhancing restoration transformers. \textbf{Third,} we introduce TEA to the vanilla Vision Transformer to design TEAFormer. Experiments show that TEA achieves State-of-The-Art (SoTA) performance in various image restoration tasks, while having faster convergence and more robust generalization.

\section{Related Work}
\label{sec:related_work}

\noindent \textbf{Vision Transformers with inductive bias.} Following the pioneering introduction of the Transformer~\cite{vaswani2017attention} into vision tasks by ViT~\cite{dosovitskiy2021an}, SA has emerged as a key component in vision networks. However, ViT is challenging to train, exhibits slow convergence, and requires substantial data support~\cite{touvron2021training}. Previous works, such as ViTAE~\cite{xu2021vitae,zhang2022vitaev2}, attributed this to a lack of inductive bias and advocated integrating inductive bias with ViT to develop more efficient ones. AACNet~\cite{bello2019attention} emphasized the importance of translation equivariance for visual tasks. The Swin~\cite{liu2021Swin} and CSwin~\cite{dong2022cswin} Transformers incorporated local processing into the vision Transformer, thus proposing window attention. CVT~\cite{wu2021cvt} and FasterViT~\cite{hatamizadehfastervit} sequentially stack convolutions and attention layers, resulting in a hybrid architecture. PVT~\cite{wang2021pyramid,wang2022pvt} used average pooling for downsampling to achieve weak translation equivariance. Both DAT~\cite{xia2022vision} and BiFormer~\cite{zhu2023biformer} dynamically determined the windows accessible to each token. Recently, some studies have introduced translation equivariance into ViT by modifying position encoding~\cite{bello2019attention,xu20232} and attention calculation logic~\cite{ramachandran2019stand,rojas2024making}.

\noindent \textbf{Recent image restoration Transformers.} Similar to the development of vision transformers, the design of image restoration models also focuses on how to better design local global information processors~\cite{liang2021swinir,Zamir2021Restormer,chen2023hat,zhou2023srformer,li2023grl}. Since the resolution of the input image is generally high in the image restoration task, directly using SA will result in a huge computation cost. To address this issue, SwinIR~\cite{liang2021swinir} introduced window attention~\cite{liu2021Swin}. HAT~\cite{chen2023hat} enlarged the receptive field of the restoration transformers with the squeeze-excitation module~\cite{hu2018squeeze} and the overlapping mechanism. Restormer~\cite{Zamir2021Restormer} and DAT~\cite{chen2023dual} performed channel-wise attention, thereby facilitating more efficient feature re-weighting. GRL~\cite{li2023grl} extracted the feature on local, regional, and global scales by integrating the attention of the anchor window. IPG~\cite{tian2024image} exploited the flexibility of the network by proposing the degree-variant graph solution, breaking the rigidity of the previous methods.


\setlength{\abovedisplayskip}{-1mm}
\setlength{\belowdisplayskip}{0pt}

\section{Translation Equivariance}
\label{sec:definition_te}

In this section, we define the translation equivariance and propose two fundamental strategies for integrating translation equivariance into restoration networks.


\noindent \textbf{Definition 3.1. Translation Equivariance.} We call function $\Phi(\cdot)$ is translation equivariant to translation operation $\mathcal{T}(\cdot)$, if $\Phi(\mathcal{T}(x)) = \mathcal{T}(\Phi(x))$, where $x$ is the input signal. 

In image restoration tasks, $x$ is an image or a feature in the latent space typically, $\mathcal{T}(\cdot)$ refers to a shift in pixels~\cite{hassani2023neighborhood}. Based on this definition, two theorems are derived:






\noindent \textbf{Theorem 3.2.}\label{Theorem32} If $\Phi(x)_i$ is transformed from $x_j=[i-b, i+b]$, where $b$ is the sliding boundary, the function $\Phi(\cdot)$ fulfills Definition 3.1.

\noindent \textbf{Theorem 3.3.} If the functions $\Phi_1(\cdot)$ and $\Phi_2(\cdot)$ fulfill Definition 3.1, the functions $\Phi_2(\Phi_1(\cdot))$ and $\Phi_1(\cdot) + \Phi_2(\cdot)$ also fulfill Definition 3.1.


Reformulating the existing operators in line with Theorem 3.2 gives them the translation equivariance property. By stacking these operators, either in parallel or in serial, as specified in Theorem 3.3, more complex architectures are also capable of translation equivariance. Consequently, the model can be further enhanced via translation equivariance.


\begin{figure}[ht]
\centering
\includegraphics[width=\linewidth]{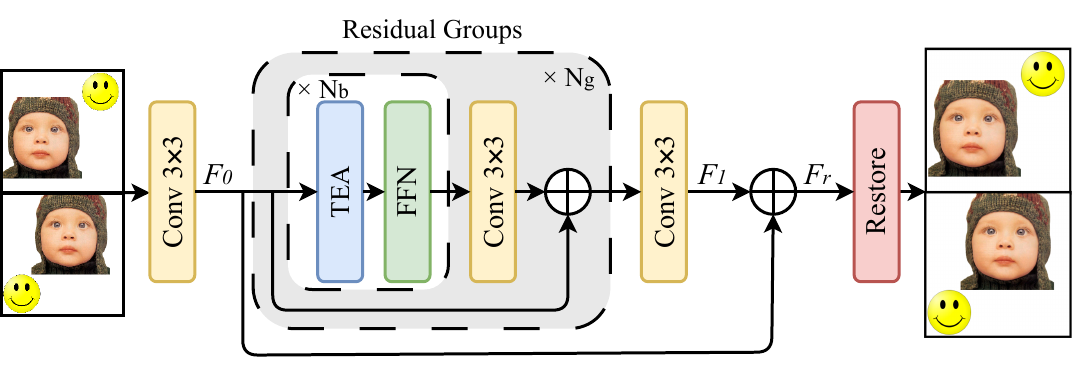}
\vspace{-5mm}
\caption{The overall architecture of our proposed TEAFormer.}\label{fig:overall}
\vspace{-5mm}
\end{figure}

\section{Method}

In this section, we conform to Theorems 3.1 and 3.2 to devise a translation equivariance restoration transformer, termed Translation Equivariance trAnsFormer (TEAFormer). We first present the overall architecture of TEAFormer and then introduce its components in detail.

\subsection{Overall Architecture}

The overall architecture of the proposed TEAFormer is illustrated in Figure~\ref{fig:overall} with image super-resolution (SR) presented as a demonstration. Given an input image $\mathit{I} \in \mathbb{R}^{H \times W \times 3}$, where $H \times W$ denotes the resolution, TEAFormer first applies a convolution layer to obtain a shallow feature $\mathit{F_0} \in \mathbb{R}^{H \times W \times D}$, where $D$ is the embedding dimension of the network. Next, following classical restoration networks~\cite{liang2021swinir}, TEAFormer adopts a residual-in-residual structure to construct a deep feature extraction module comprised of $N_g$ Translation Equivariance Groups (TEGs). Each TEG consists of $N_b$ Translation Equivariance Blocks (TEBs) and a convolution layer. Each TEB, in turn, is composed of one Translation Equivariance Attention (TEA) and one feed-forward network. Subsequently, a convolutional layer is employed to extract the deep feature $\mathit{F_1} \in \mathbb{R}^{H \times W \times D}$ from the feature extracted by TEGs. After extraction through the deep feature extraction module, we use the restore module to generate the high-quality image $\mathit{\hat{I}}$ from the feature $\mathit{F_r} = \mathit{F_0} + \mathit{F_1}$. As for other image restoration tasks that do not involve changes in resolution, we build our TEAFormer following Restormer~\cite{Zamir2021Restormer}.

\subsection{Adaptive Translation Equivariance}\label{sec:tea}

\begin{figure}[ht]
\centering
\includegraphics[width=\linewidth]{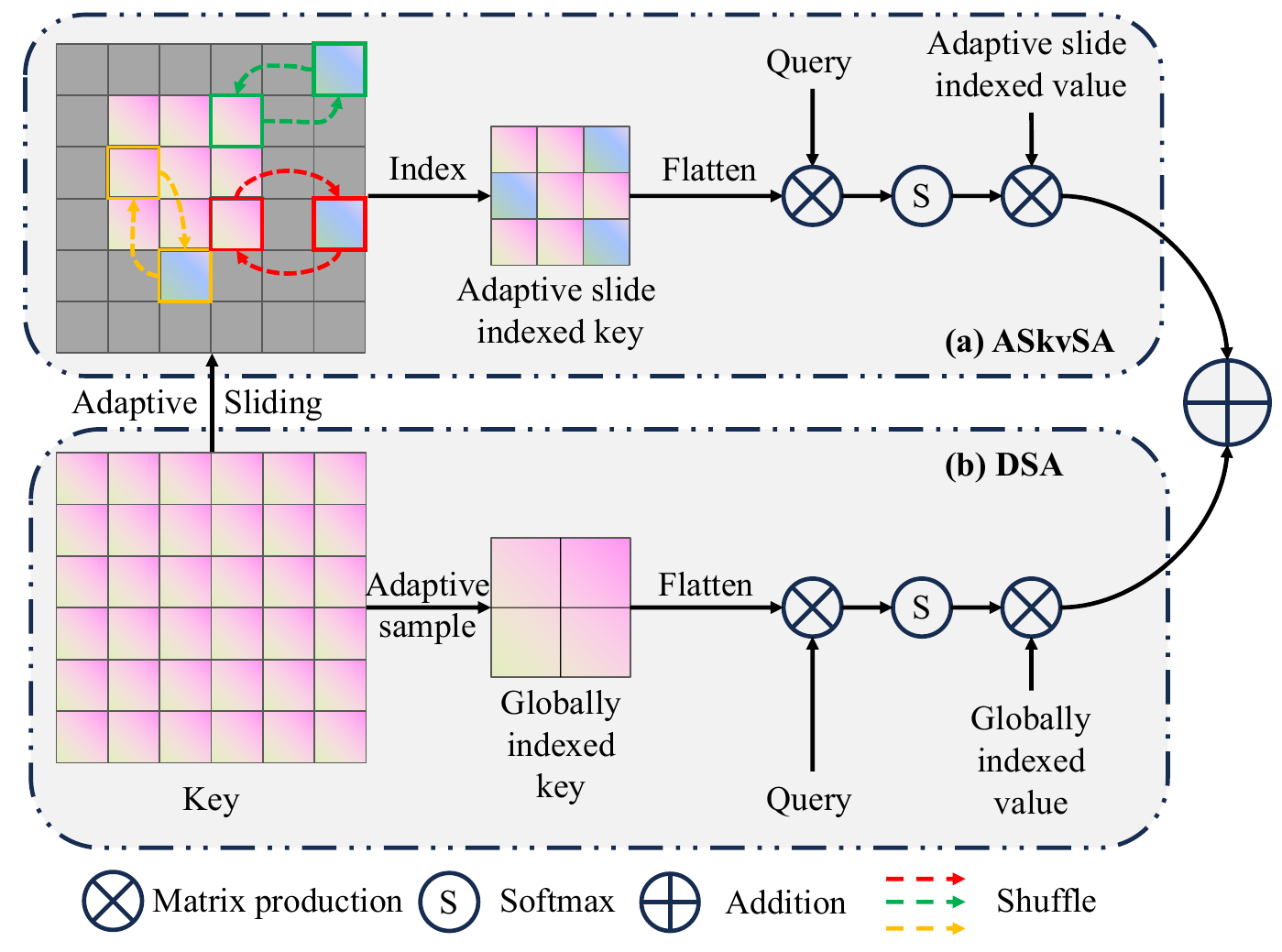}
\vspace{-6mm}
\caption{The inner structure of TEA with two components. (a) Adaptive Sliding key-value Self Attention (ASkvSA) and (b) Downsampled Self Attention (DSA).}\label{fig:overall_tea}
\vspace{-6mm}
\end{figure}

Adaptive Translation Equivariance is integrated through the Adaptive Sliding Key-Value Self Attention (ASkvSA) and Downsampled Self Attention (DSA) as shown in Figure~\ref{fig:overall_tea}. 

\noindent \textbf{Key-value indexing in self-attention.} The nature of Self Attention (SA) without position bias lies in the indexing of key-value pairs, followed by feature re-weighting. Given a flattened input $X \in \mathbb{R}^{N \times D}$, where $N$ represents the number of tokens and $D$ represents the embedding dimension, self-attention can be formulated as following: 

\begin{equation}\label{eq:vsa}
\text{SA}(X) = \text{Softmax} (\frac{XW_q (XW_k)^{\top}}{\sqrt{D}}) XW_v,
\end{equation}
\noindent where $W_q, W_k, \ \text{and} \ W_v$ denote the weights of the linear projections utilized to compute the query ($Q$), key ($K$), and value ($V$) from input $X$.

In Eq.~\ref{eq:vsa}, the softmax similarity function is utilized for the generation of attention maps between a query and a key, while the dot product is employed for re-weighting between the attention map and a value. Given that subscripting a matrix with $i$ returns the $i$-th row as a vector, an alternative equation of self-attention can be written as follows.

\begin{equation}\label{eq:vsa_index}
\text{SA}_{i}(X) = \frac{\sum_{j=1}^N \text{sim}{(Q_i, K_j)} V_j}{\sum_{j=1}^N \text{sim}{(Q_i, K_j)}},
\end{equation}

\noindent where $\text{sim}(Q_i, K_j)=\text{exp}(\frac{{Q_i}^{\top} {K_j}}{\sqrt{D}})$. 

Eq.~\ref{eq:vsa_index} illustrates that self-attention indexes key-value pairs ($K_j \ \text{and} \ V_j, \ j = 1, \cdots, N$) for each query ($Q_i$). 

\noindent \textbf{Sliding key-value Self Attention.} In accordance with Theorem 3.2, we demonstrate that sliding window attention constitutes a generalized form of self-attention, which can be designated as Sliding Key-Value Self-Attention (\textbf{SkvSA}). SkvSA is formulated as follows.


\begin{equation}\label{eq:skvsa}
\text{SkvSA}_{i}(X) = \frac{\sum_{j=(i - b_1) \ \text{mod} \ s = 0}^{(i + b_2) \ \text{mod} \ s = 0} \text{sim}{(Q_i, K_j)} V_j}{\sum_{j=(i - b_1) \ \text{mod} \ s = 0}^{(i + b_2) \ \text{mod} \ s = 0} \text{sim}{(Q_i, K_j)}},
\end{equation}

\noindent where $b_1$ and $b_2$ are the sliding boundary on the left and right when sliding on the one-dimensional vector $X$. In our SkvSA, $b_1=b_2=\frac{w\cdot s}{2}$, where $w$ is the size of the sliding window and $s$ is the stride size during sliding.

Eq.~\ref{eq:skvsa} says that, for the $i$-th query $Q_i$, SkvSA systematically extracts the key-value pairs from a designated sliding window, extending from the index $i - \frac{w\cdot s}{2}$ to the index $i + \frac{w\cdot s}{2}$ with stride $s$ \footnote{We set $w \cdot s \le N$ in experiments thus enabling SkvSA to slide on $X$.}. Subsequently, SkvSA calculates the attention map and re-weights the feature in place. 

\noindent \textbf{Boundary processing.} Note that, in scenarios where $i - \frac{w\cdot s}{2} < 0$ or $i + \frac{w\cdot s}{2} > N$ occur, the slide indices of the key-value pairs $j$ will exceed the boundary of the input $X$, leading to an indexing failure. To ameliorate the issue associated with slide indexing at sequence's boundaries, we propose a boundary blocking indexing method. Specifically, for queries located at the boundary of the sequence, the slided indices of key-value pairs are ``blocked" and converted into indexing within a predefined blocking window. In this study, the sizes of the predefined blocking window and the sliding window are identical. Accordingly, the sliding boundaries $b$ can be appropriately revised as follows:

\vspace{-5pt}
\begin{equation}
b = (b_1,b_2) = \left\{
\begin{aligned}
&(0,\frac{w\cdot s}{2}), &i - \frac{w\cdot s}{2} < 0, \\
&(\frac{w\cdot s}{2},N), &i + \frac{w\cdot s}{2} > N, \\
&(\frac{w\cdot s}{2},\frac{w\cdot s}{2}), &\text{others}.
\end{aligned}
\right.
\end{equation}

The SkvSA and its boundary processing are capable of extracting and processing information via a fixed sliding window, adhering to the translation equivariance property as established in Theorem 3.2.


\noindent \textbf{Adaptive Sliding key-value Self Attention.} However, the fixed window size constrains the receptive field, leading to the entrapment of restoration transformers within rigidity~\cite{tian2024image}. To mitigate this rigidity, it is anticipated that the SkvSA will ascertain the optimal window size and position specific to each query $Q_i$. To achieve this endeavor, we devise the Adaptive Sliding key-value Self Attention (\textbf{ASkvSA}), which adaptively determines key-value indices and shuffles them within the fixed sliding window of $Q_i$.


As shown in Figure~\ref{fig:overall_tea} (a), with regard to Key $K$ and Value $V$, ASkvSA initially reshapes $K$ and $V$ from one-dimensional (1d) vectors into two-dimensional (2d) matrices. Subsequently, it applies a 2d depth-wise convolution with kernel size $k$ to generate their adaptive indices $\mathcal{F} \in \mathbb{R}^{H \times W \times 2}$. $\mathcal{F}[{h,w}]$ is a vector of length 2 in the $h-$th row and $w-$th column of $\mathcal{F}$, which can describe a coordinate. Following this, ASkvSA performs key-value indexing based on established indices $\mathcal{F}_K$ and $\mathcal{F}_V$ by shuffling the pixel located at the coordinate $\mathcal{F}[{h,w}]$ of $K$ and $V$ to the pixel at the coordinate ($h,w$) of $K$ and $V$. 

\begin{equation}\label{eq:askvsa}
\text{ASkvSA}(X) = \frac{\sum_{j=\mathcal{F}(K),l=\mathcal{F}(V)} \text{sim}{(Q_i, K_j)} V_l}{\sum_{j=\mathcal{F}(K)} \text{sim}{(Q_i, K_j)}}.
\end{equation}

Adaptive slide indexing does not compromise the translation equivariance of ASkvSA. ASkvSA stacks convolutions with SkvSA in serial to generate adaptive indices of key-value pairs. Given that convolution satisfies the property of translation equivariance, when $K_{h,w}$ is translated to $K_{h+\delta_h,w+\delta_w}$, key's adaptive indices $\mathcal{F}_K$ will shuffle the pixel located at $\mathcal{F}[{h+\delta_h,w+\delta_w}]$ of $K$ to the pixel at ($h+\delta_h,w+\delta_w$) of $K$. According to Theorem 3.3, ASkvSA satisfies the translation equivariance property.

\noindent \textbf{Adaptively aggregated with global key-value pairs.} Although ASkvSA can identify appropriate key-value pairs beyond a fixed sliding window, some remote but valid pixels still escape ASkvSA's indexing. Consequently, we propose downsampled self-attention (\textbf{DSA}), which adaptively combines global key value indexes through downsampling, providing the network with a coarse, yet efficient global indexing of key-value pairs. DSA can be formulated as:

\vspace{-3mm}
\begin{equation}\label{eq:dsa}
\text{DSA}_{i}(X) = \frac{\sum_{j=1}^{N_{d}} \text{sim}{(Q_i, K'_j)} V'_j}{\sum_{j=1}^{N_{d}} \text{sim}{(Q_i, K'_j)}},
\end{equation}
\vspace{-3mm}

\noindent where $K' \ \text{and} \ V'$ represents the downsampled key-value pair calculated in ASkvSA, and $N_d$ denotes the number of tokens associated with the downsampled key-value pair. We employ average pooling as the downsampling operator due to its efficiency and coarse fulfillment~\footnote{~\cite{rojas2022learnable} proved that downsampling can meet strict TE through learnable polyphase (LP). We compare these downsamplers in Sec.~\ref{sec:ablation}} of TE.

Finally, according to Figure~\ref{fig:overall_tea}, the ASkvSA and DSA outputs are adaptively aggregated to obtain our TEA.

\begin{equation}\label{eq:tea}
\text{TEA}_{i}(X) = \alpha_{s} \ \text{ASkvSA}(X) + \alpha_{d} \ \text{DSA}(X),
\end{equation}

\noindent where $\alpha_{s}$ and $\alpha_{d}$ are learnable parameters. TEA stacks DSA with ASkvSA in parallel. According to Theorem 3.3, TEA satisfies the translation equivariance property.

\subsection{Computational Cost and Hyper-Parameters}\label{sec:stacking}

We demonstrate that TEA can maintain a computational cost that is similar to that of window attention~\cite{liu2021Swin} under well-designed hyperparameter settings. We first list four integral hyperparameters of TEA:

\begin{itemize}
\item $w$: the sliding window size of ASkvSA;
\item $s$: the sliding stride size of ASkvSA;
\item $k$: the kernel size of adaptive indexes generation;
\item $N_d$: the number of tokens in DSA.
\end{itemize}

According to Section~\ref{sec:tea}, TEA uses $3ND^2$ FLOPs for the Query-Key-Value linear projections, $2NDk^2$ FLOPs for the depth-wise convolution in adaptive indices generation, $Nw^2D$ FLOPs for attention map computation between query $Q \in \mathbb{R}^{N \times D}$ and indexed key $K \in \mathbb{R}^{w^2 \times D}$, $Nw^2D$ FLOPs for feature re-weighting between attention map and indexed value $V \in \mathbb{R}^{w^2 \times D}$, and $2NN_{d}D$ FLOPs for DSA. The total FLOPs of TEA is $3ND^2 + 2NDk^2 + 2Nw^2D + 2NN_{d}D$. With the hyper-parameters held constant, the FLOPs of TEA exhibit linear growth as a function of $N$, thereby implying that TEA is $\mathcal{O}(N)$.

Consequently, when the hyper-parameters of TEA are adjusted carefully, its computational cost becomes comparable to that of window attention. In our TEAFormer, the hyper-parameter $w$ has been set to 15, $k$ to 3, and $N_d$ to 16, thereby modulating the computational complexity of TEA to be even slightly lower than that of window attention with window size 16.


\begin{table*}[!ht]
\centering
\scalebox{0.75}{
\begin{tabular}{c|c|c|c|c|c|c|c|c|c|cccc}
\toprule[0.15em]
\multirow{3}{*}{\textbf{Model}} & \multirow{3}{*}{\textbf{window size}} & 
\multicolumn{3}{c}{\textbf{Translation equivariance}} &
\textbf{Params} & 
\textbf{FLOPs} &
\textbf{Latency} &
\textbf{Performance} &
\textbf{NTK~\cite{jacot2018neural}} & 
\textbf{SRGA~\cite{liu2023evaluating}} 
\\
\specialrule{0em}{1pt}{1pt}
\cline{3-11}
\specialrule{0em}{1pt}{1pt}
&  & \multirow{2}{*}{SkvSA} & \multirow{2}{*}{ASkvSA} & \multirow{2}{*}{DSA} & \multirow{2}{*}{(M)} & \multirow{2}{*}{(T)} & \multirow{2}{*}{(ms)} & Urban100~\cite{Urban100} & \multirow{2}{*}{Condition $\downarrow$} & \multirow{2}{*}{Value  $\downarrow$} \\
&  &  &  &  & & & & PSNR$\uparrow$ / SSIM$\uparrow$ & \\
\midrule[0.15em]
SwinIR~\cite{liang2021swinir} & 8 &  &  &  & 11.90 & 0.462 & 130.0 & 27.45 / 0.8254 & 1746.49 & 3.655 \\
SwinIR-Large & 16 &  &  &  & 21.42 & 0.897 & 214.5 & 27.94 / 0.8362 & 1554.65 & 3.610 \\
TEAFormer & 15 & \ding{52} &  &  & 21.86 & 0.688 & 230.1 & 28.31 / 0.8444 & 243.75 & 3.206 \\
TEAFormer & 15 &  & \ding{52} &  & 21.98 & 0.757 & 284.4 & 28.47 / 0.8457 & 283.99 & 3.298 \\
TEAFormer & 15 & \ding{52} &  & \ding{52} & 21.86 & 0.966 & 340.9 & 28.49 / 0.8470 & 203.06 & 3.261 \\
TEAFormer & 15 &  & \ding{52} & \ding{52} & 21.98 & 1.035 & 386.7 & 28.67 / 0.8489 & 236.78 & 3.275 \\
\bottomrule[0.15em]
\end{tabular}
}
\vspace{-3mm}
\caption{Ablation study on TEA. Our TEA enhances performance, convergence speed, and generalization.}\label{tab:ablation_tea}
\end{table*}

\begin{table*}[!ht]
\footnotesize
\setlength\tabcolsep{5pt}
\centering
\normalsize
\scalebox{0.75}{
\begin{tabular}{c|c|c|c|c|c|c|c|c|c|c|c|c|c|c}
\toprule[0.15em]
\multirow{2}{*}{Method} & \multirow{2}{*}{Scale} & Params & FLOPs & Latency & \multicolumn{2}{c|}{Set5~\cite{Set5}} & \multicolumn{2}{c|}{Set14~\cite{Set14}} &  \multicolumn{2}{c|}{B100~\cite{BSD100}} &  \multicolumn{2}{c|}{Urban100~\cite{Urban100}} &  \multicolumn{2}{c}{Manga109~\cite{Manga109}}  
\\
\cline{4-15}
& & (M) & (T) & (ms) & PSNR $\uparrow$ & SSIM $\uparrow$ & PSNR $\uparrow$ & SSIM $\uparrow$ & PSNR $\uparrow$ & SSIM $\uparrow$ & PSNR $\uparrow$ & SSIM $\uparrow$ & PSNR $\uparrow$ & SSIM $\uparrow$
\\
\midrule[0.15em]
EDSR~\cite{lim2017enhanced} & & 40.7 & 6.006 & 147.1 
& {38.11} & {0.9602}
& {33.92} & {0.9195}
& {32.32} & {0.9013}
& {32.93} & {0.9351}
& {39.10} & {0.9773}
\\
RCAN~\cite{zhang2018rcan} & & 15.4 & 2.259 & 155.5
& {38.27} & {0.9614}
& {34.12} & {0.9216}
& {32.41} & {0.9027}
& {33.34} & {0.9384}
& {39.44} & {0.9786}
\\
\hdashline
SwinIR-L~\cite{liang2021swinir} &  & 0.91 & 156.4 & 188.6
& {38.14} & {0.9611}
& {33.86} & {0.9206}
& {32.31} & {0.9012}
& {32.76} & {0.9340}
& {39.12} & {0.9783}
\\
SwinIR~\cite{liang2021swinir} &  & 11.8 & 1.848 & 514.4
& {38.42} & {0.9623}
& {34.46} & {0.9250}
& {32.53} & {0.9041}
& {33.81} & {0.9427}
& {39.92} & {0.9797}
\\
EDT~\cite{ijcai2023p121} &  & 11.5 & 1.965 & 1010
& {38.45} & {0.9624}
& {34.57} & {0.9258}
& {32.52} & {0.9041}
& {33.80} & {0.9425}
& {39.93} & {0.9800}
\\
SRFormer~\cite{zhou2023srformer} & {$\times$2} & 10.3 & 1.790 & 787.2 
& {38.51} & {0.9627}
& {34.44} & {0.9253}
& {32.57} & {0.9046}
& {34.09} & {0.9449}
& {40.07} & {0.9802}
\\
HAT~\cite{chen2023hat} & & 20.6 & 3.662 & 901.9
& {38.63} & {0.9630}
& \textcolor{blue}{34.86} & \textcolor{blue}{0.9274}
& \textcolor{blue}{32.62} & \textcolor{blue}{0.9053}
& {34.45} & \textcolor{blue}{0.9466}
& \textcolor{blue}{40.26} & \textcolor{blue}{0.9809}
\\
GRL-B*~\cite{li2023grl} & & 20.1 & 4.422 & 2621
& 38.48 & 0.9627 
& 34.64 & 0.9265 
& 32.55 & 0.9045
& 33.97 & 0.9437 
& 40.06 & 0.9804
\\
DAT~\cite{chen2023dual} & & 14.7 & 2.155 & 1215
& {38.58} & {0.9629}
& {34.81} & {0.9272}
& {32.61} & {0.9051}
& {34.37} & {0.9458}
& {40.33} & {0.9807}
\\
IPG~\cite{tian2024image} & & 16.8 & 4.732 & 2703
& \textcolor{blue}{38.61} & \textcolor{red}{0.9632} 
& 34.73 & 0.9270 
& 32.60 & 0.9052 
& \textcolor{blue}{34.48} & 0.9464 
& 40.24 & \textcolor{red}{0.9810}
\\
\textbf{TEAFormer-L~(Ours)} & 
& \textbf{0.83} & \textbf{0.237} & \textbf{254.0}
& \textbf{{38.27}} & \textbf{{0.8618}}
& \textbf{{34.26}} & \textbf{{0.9240}}
& \textbf{{32.43}} & \textbf{{0.9029}}
& \textbf{{33.51}} & \textbf{{0.9396}}
& \textbf{{39.56}} & \textbf{{0.9792}}
\\
\textbf{TEAFormer~(Ours)} & 
& \textbf{21.8} & \textbf{4.133} & \textbf{1493}
& \textbf{\textcolor{red}{38.64}} & \textbf{\textcolor{blue}{0.9631}}
& \textbf{\textcolor{red}{35.12}} & \textbf{\textcolor{red}{0.9284}}
& \textbf{\textcolor{red}{32.72}} & \textbf{\textcolor{red}{0.9064}}
& \textbf{\textcolor{red}{35.23}} & \textbf{\textcolor{red}{0.9509}}
& \textbf{\textcolor{red}{40.32}} & \textbf{{0.9807}}
\\
\midrule
EDSR~\cite{lim2017enhanced} & & 43.1 & 1.853 & 49.13
& {32.46} & {0.8968}
& {28.80} & {0.7876}
& {27.71} & {0.7420}
& {26.64} & {0.8033}
& {31.02} & {0.9148}
\\
RCAN~\cite{zhang2018rcan} & & 15.6 & 0.587 & 46.15
& {32.63} & {0.9002}
& {28.87} & {0.7889}
& {27.77} & {0.7436}
& {26.82} & {0.8087}
& {31.22} & {0.9173}
\\
\hdashline
SwinIR-L~\cite{liang2021swinir} &  & 0.88 & 0.039 & 48.71
& 32.44 & 0.8976 
& 28.77 & 0.7858 
& 27.69 & 0.7406 
& 26.47 & 0.7980 
& 30.92 & 0.9151
\\
SwinIR~\cite{liang2021swinir} &  & 11.9 & 0.462 & 130.0
& {32.92} & {0.9044}
& {29.09} & {0.7950}
& {27.92} & {0.7489}
& {27.45} & {0.8254}
& {32.03} & {0.9260}
\\
EDT~\cite{ijcai2023p121} &  & 11.6 & 0.514 & 251.3
& 32.82 & 0.9031 
& 29.09 & 0.7939 
& 27.91 & 0.7483 
& 27.46 & 0.8246 
& 32.03 & 0.9254 
\\
SRFormer~\cite{zhou2023srformer} & {$\times$4} & 10.4 & 0.470 & 189.2
& {32.93} & {0.9041}
& {29.08} & {0.7953}
& {27.94} & {0.7502}
& {27.68} & {0.8311}
& {32.44} & {0.9271}
\\
HAT~\cite{chen2023hat} & & 20.8 & 0.938 & 227.7
& {33.04} & {0.9056}
& {29.23} & {\textcolor{blue}{0.7973}}
& \textcolor{blue}{28.00} & {0.7517}
& {27.97} & {0.8368}
& {32.48} & {0.9292}
\\
GRL-B*~\cite{li2023grl} & 
& 20.2 & 1.129 & 761.8
& 32.90 & 0.9039
& 29.14 & 0.7956
& 27.96 & 0.7497 
& 27.53 & 0.8276 
& 32.19 & 0.9266
\\
DAT~\cite{chen2023dual} & 
& 14.8 & 0.561 & 296.8
& {33.08} & {0.9055}
& {29.23} & {\textcolor{blue}{0.7973}}
& \textcolor{blue}{28.00} & {0.7515}
& {27.87} & {0.8343}
& {32.51} & {0.9291}
\\
IPG~\cite{tian2024image} & 
& 17.0 & 1.206 & 646.7
& \textcolor{blue}{33.15} & \textcolor{blue}{0.9062} 
& \textcolor{blue}{29.24} & \textcolor{blue}{0.7973} 
& 27.99 & \textcolor{blue}{0.7519} 
& \textcolor{blue}{28.13} & \textcolor{blue}{0.8392} 
& \textcolor{blue}{32.53} & \textcolor{blue}{0.9300}
\\
\textbf{TEAFormer-L~(Ours)} & 
& \textbf{0.83} & \textbf{0.059} & \textbf{69.64}
& \textbf{{32.63}} & \textbf{{0.9002}}
& \textbf{{28.96}} & \textbf{{0.7899}}
& \textbf{{27.82}} & \textbf{{0.7442}}
& \textbf{{27.04}} & \textbf{{0.8105}}
& \textbf{{31.56}} & \textbf{{0.9197}}
\\
\textbf{TEAFormer~(Ours)} & 
& \textbf{22.0} & \textbf{1.035} & \textbf{386.7}
& \textbf{\textcolor{red}{33.23}} & \textbf{\textcolor{red}{0.9066}}
& \textbf{\textcolor{red}{29.46}} & \textbf{\textcolor{red}{0.8009}}
& \textbf{\textcolor{red}{28.11}} & \textbf{\textcolor{red}{0.7544}}
& \textbf{\textcolor{red}{28.67}} & \textbf{\textcolor{red}{0.8489}}
& \textbf{\textcolor{red}{32.99}} & \textbf{\textcolor{red}{0.9323}}
\\
\bottomrule[0.15em]           
\end{tabular}
}
\vspace{-3mm}
\caption{\textit{\textbf{Classical image SR}} results. Performance and complexity are shown for better comparison. Dash lines separate the results of CNNs and Transformers. The best two scores in each column are highlighted in \textcolor{red}{red} and \textcolor{blue}{blue}. Methods with ``*" are replicated with standard training setting, following IPG~\cite{tian2024image}. The ``-L" indicates that the given method is a lightweight version.}\label{tab:super-resolution}
\end{table*}

\begin{figure*}[!ht]
\centering
\newcommand{\h}{0.105}
\newcommand{\wa}{0.12}
\newcommand{\wb}{0.16}
\newcommand{\g}{-0.7mm}
\renewcommand{\tabcolsep}{1.8pt}
\renewcommand{\arraystretch}{1}
\resizebox{1.00\linewidth}{!} {
    \begin{tabular}{l}
        \normalsize
        \newcommand{\name}{images/csr/}
        \renewcommand{\h}{0.086}
        \newcommand{\wq}{0.180}
        \begin{tabular}{cc}
        \normalsize
            \begin{adjustbox}{valign=t}
                \begin{tabular}{c}
                    \includegraphics[height=0.205\textwidth, width=0.374\textwidth]{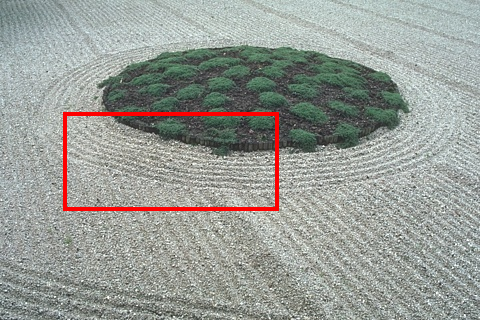}
                    \\
                    BSD100 ($4\times$): img\_86016
                \end{tabular}
            \end{adjustbox}
            \begin{adjustbox}{valign=t}
                \begin{tabular}{ccccc}
                    \includegraphics[height=\h \textwidth, width=\wq \textwidth]{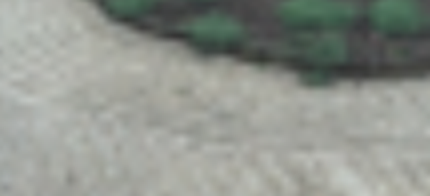} \hspace{\g} &
                    \includegraphics[height=\h \textwidth, width=\wq \textwidth]{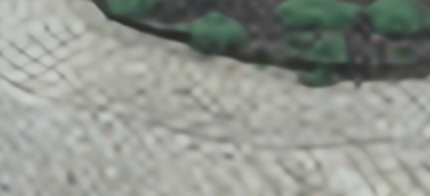} \hspace{\g} &
                    \includegraphics[height=\h \textwidth, width=\wq \textwidth]{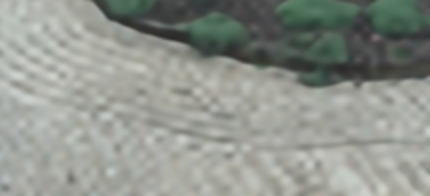} \hspace{\g} &
                    \includegraphics[height=\h \textwidth, width=\wq \textwidth]{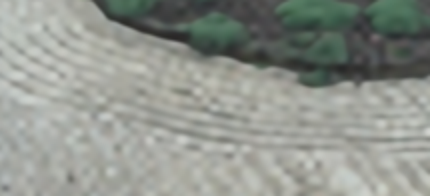} \hspace{\g} 
                    \\
                    Bicubic \hspace{\g} &
                    RCAN~\cite{zhang2018rcan} \hspace{\g} &
                    SwinIR~\cite{liang2021swinir} \hspace{\g} &
                    HAT~\cite{chen2023hat} \hspace{\g} 
                    \\
                    \vspace{-4.0mm}
                    \\
                    
                    \includegraphics[height=\h \textwidth, width=\wq \textwidth]{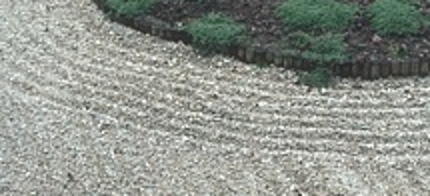} \hspace{\g} &
                    \includegraphics[height=\h \textwidth, width=\wq \textwidth]{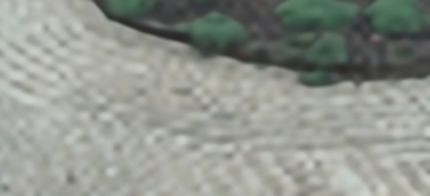} \hspace{\g} &
                    \includegraphics[height=\h \textwidth, width=\wq \textwidth]{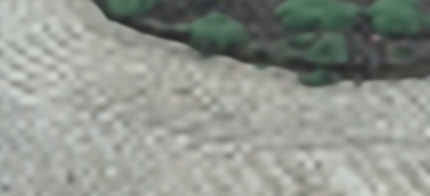} \hspace{\g} &
                    \includegraphics[height=\h \textwidth, width=\wq \textwidth]{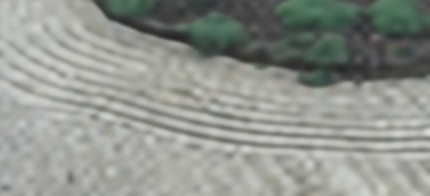} \hspace{\g} 
                    \\ 
                    HQ &
                    SRFormer~\cite{zhou2023srformer} &
                    IPG~\cite{tian2024image} &
                    \textbf{TEAFormer (Ours)}  \\
                \end{tabular}
            \end{adjustbox}
        \end{tabular} \\
    \end{tabular}
}
\resizebox{1.00\linewidth}{!} {
    \begin{tabular}{l}
        \normalsize
        \newcommand{\name}{images/csr2/}
        \renewcommand{\h}{0.086}
        \newcommand{\wq}{0.180}
        \begin{tabular}{cc}
        \normalsize
            \begin{adjustbox}{valign=t}
                \begin{tabular}{c}
                    \includegraphics[height=0.205\textwidth, width=0.374\textwidth]{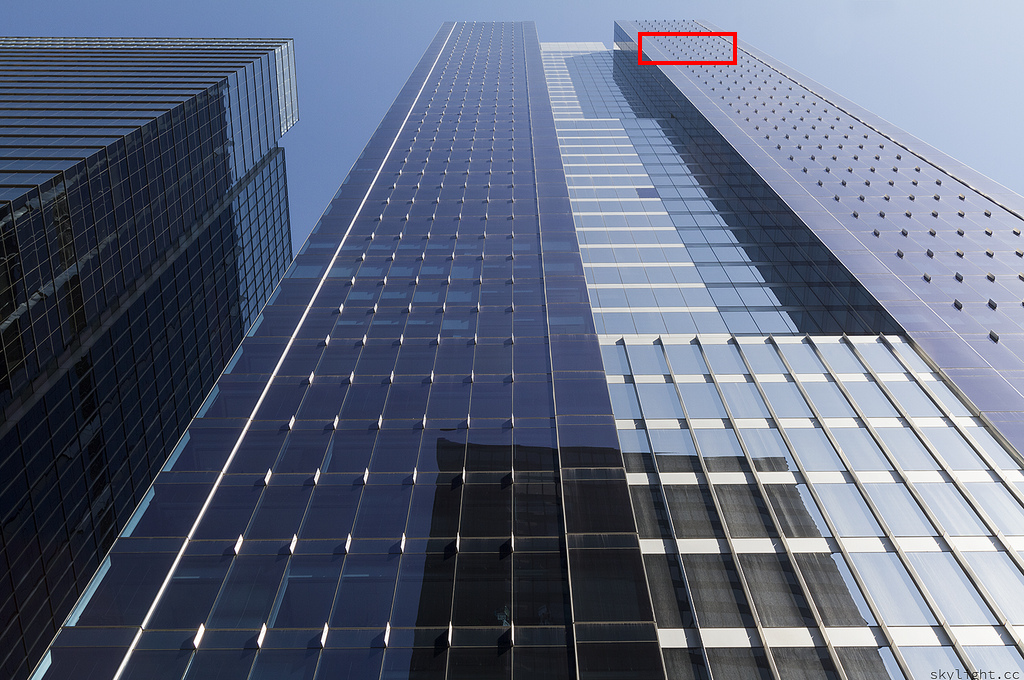}
                    \\
                    Urban100 ($4\times$): img\_033
                \end{tabular}
            \end{adjustbox}
            \begin{adjustbox}{valign=t}
                \begin{tabular}{ccccc}
                    \includegraphics[height=\h \textwidth, width=\wq \textwidth]{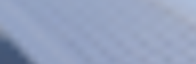} \hspace{\g} &
                    \includegraphics[height=\h \textwidth, width=\wq \textwidth]{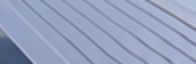} \hspace{\g} &
                    \includegraphics[height=\h \textwidth, width=\wq \textwidth]{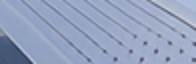} \hspace{\g} &
                    \includegraphics[height=\h \textwidth, width=\wq \textwidth]{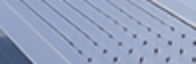} \hspace{\g} 
                    \\
                    Bicubic \hspace{\g} &
                    RCAN~\cite{zhang2018rcan} \hspace{\g} &
                    SwinIR~\cite{liang2021swinir} \hspace{\g} &
                    HAT~\cite{chen2023hat} \hspace{\g} 
                    \\
                    \vspace{-4.0mm}
                    \\
                    
                    \includegraphics[height=\h \textwidth, width=\wq \textwidth]{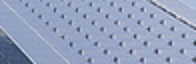} \hspace{\g} &
                    \includegraphics[height=\h \textwidth, width=\wq \textwidth]{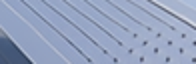} \hspace{\g} &
                    \includegraphics[height=\h \textwidth, width=\wq \textwidth]{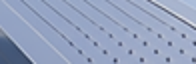} \hspace{\g} &
                    \includegraphics[height=\h \textwidth, width=\wq \textwidth]{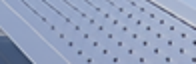} \hspace{\g} 
                    \\ 
                    HQ &
                    SRFormer~\cite{zhou2023srformer} &
                    IPG~\cite{tian2024image} &
                    \textbf{TEAFormer (Ours)}  \\
                \end{tabular}
            \end{adjustbox}
        \end{tabular} \\
    \end{tabular}
}
\vspace{-4mm}
\caption{Qualitative comparison with recent SOTA methods on the \textit{\textbf{image SR}} task.} \label{fig:sr}
\end{figure*}

\section{Experiments}


\subsection{Ablation study}

We analyze the influence exerted by the TEA on both performance, computational complexity, generalization, and training convergence in Table~\ref{tab:ablation_tea}. In this experiment, we disassemble TEA into several plug-and-play modules and gradually add them into the SwinIR baseline restoration network. To ensure comparable parameters and FLOPs among the models being evaluated, we adjusted the window size of SwinIR to 16 and increased the number of parameters to 21 M, which is termed \textbf{SwinIR-Large}. Following the incorporation of our SkvSA, there is a marked improvement in the convergence rate and generalization abilities. Subsequent to the integration of our ASkvSA and DSA, there is a progressive enhancement in the model's performance, without detrimental effects on convergence and generalization. The results of this experiment demonstrate the efficacy of each component within the TEA. Furthermore, it illustrates that TEA resolves the dilemma between self-attention and sliding window attention, thereby attaining high performance by the global receptive field and reducing computational complexity simultaneously.

\subsection{Image SR}\label{sec:sr}


\noindent \textbf{Training settings.} We used the same training hyperparameters and evaluation methods as IPG~\cite{tian2024image}. All of our models were trained from scratch on the DF2K~\cite{DIV2K,lim2017enhanced} datasets with $64\times64$ resolution. The optimizer was Adam~\cite{kingma2014adam}, the learning rate was $2\times10^{-4}$ while using the cosine scheduler. We used $\text{L}_1$ loss as a criterion.

\begin{table*}[!ht]
\centering
\vspace{-2mm}
\setlength{\tabcolsep}{5pt}
\scalebox{0.75}{
\begin{tabular}{l | c | c c c c | c c c c | c c c c }
\toprule[0.15em]
\multirow{2}{*}{\textbf{Method}} & \textbf{Params} & \multicolumn{4}{c|}{\textbf{Indoor Scenes}} & \multicolumn{4}{c|}{\textbf{Outdoor Scenes}} & \multicolumn{4}{c}{\textbf{Combined}} \\
\cline{2-14}
& (M) & PSNR~$\textcolor{black}{\uparrow}$ & SSIM~$\textcolor{black}{\uparrow}$& MAE~$\textcolor{black}{\downarrow}$ & LPIPS~$\textcolor{black}{\downarrow}$  & PSNR~$\textcolor{black}{\uparrow}$ & SSIM~$\textcolor{black}{\uparrow}$& MAE~$\textcolor{black}{\downarrow}$ & LPIPS~$\textcolor{black}{\downarrow}$  & PSNR~$\textcolor{black}{\uparrow}$ & SSIM~$\textcolor{black}{\uparrow}$& MAE~$\textcolor{black}{\downarrow}$ & LPIPS~$\textcolor{black}{\downarrow}$   \\
\midrule[0.15em]
DPDNet$_S$~\cite{abuolaim2020defocus} & 32.3 &26.54 & 0.816 & 0.031 & 0.239 & 22.25 & 0.682 & 0.056 & 0.313 & 24.34 & 0.747 & 0.044 & 0.277\\
KPAC$_S$~\cite{Son_2021_ICCV} & 2.06 & 27.97 & 0.852 & 0.026 & 0.182 & 22.62 & 0.701 & 0.053 & 0.269 & 25.22 & 0.774 & 0.040 & 0.227 \\
IFAN$_S$~\cite{Lee2021IFAN} & 10.5 & {28.11}  & {0.861}  & {0.026} & {0.179}  & {22.76}  & {0.720} & {0.052}  & {0.254}  & {25.37} & {0.789} & {0.039} & {0.217}\\
{Restormer}$_S$~\cite{Zamir2021Restormer} & 26.1 & {28.87}  & {0.882}  & {0.025} & \textcolor{blue}{0.145} & {23.24}  & {0.743}  & \textcolor{red}{0.050} & \textcolor{blue}{0.209}  & {25.98}  & {0.811}  & \textcolor{blue}{0.038}  & \textcolor{blue}{0.178}   \\
SFNet$_S$~\cite{cui2022selective} & 13.2 & 29.16 & 0.878 & \textcolor{red}{0.023} & 0.168 & 23.45 & 0.747 & \textcolor{blue}{0.049} & 0.244 & 26.23 & 0.811 & \textcolor{red}{0.037} & 0.207 \\
{GRL}$_S$-B~\cite{li2023grl} & 19.9 & {29.06}  & \textcolor{blue}{0.886}  & \textcolor{blue}{0.024} & \textcolor{red}{0.139} & {23.45}  & \textcolor{blue}{0.761}  & \textcolor{blue}{0.049} & \textcolor{red}{0.196}  & {26.18}  & \textcolor{blue}{0.822}  & \textcolor{red}{0.037}  & \textcolor{red}{0.168} \\
{IRNeXT}$_S$~\cite{IRNeXt} & 13.2 & \textcolor{blue}{29.22} & 0.879 & \textcolor{blue}{0.024} & 0.167 & \textcolor{blue}{23.53} & 0.752 & \textcolor{blue}{0.049} & 0.244 & \textcolor{blue}{26.30} & 0.814 & \textcolor{red}{0.037} & 0.206 \\
\textbf{TEAFormer}$_S$ \textbf{(Ours)} & \textbf{15.4} & \textbf{\textcolor{red}{29.50}}  & \textbf{\textcolor{red}{0.892}}  & \textbf{\textcolor{red}{0.023}} & \textbf{0.150} & \textbf{\textcolor{red}{23.55}}  & \textbf{\textcolor{red}{0.767}}  & \textbf{\textcolor{blue}{0.050}} & \textbf{0.211}  & \textbf{\textcolor{red}{26.45}}  & \textbf{\textcolor{red}{0.828}}  & \textbf{\textcolor{red}{0.037}}  & \textbf{0.181}   \\
\midrule[0.1em]
\midrule[0.1em]
DPDNet$_D$~\cite{abuolaim2020defocus} & 32.3 & 27.48 & 0.849 & 0.029 & 0.189 & 22.90 & 0.726 & 0.052 & 0.255 & 25.13 & 0.786 & 0.041 & 0.223 \\
RDPD$_D$~\cite{abuolaim2021learning} & 8.20 & 28.10 & 0.843 & 0.027 & 0.210 & 22.82 & 0.704 & 0.053 & 0.298 & 25.39 & 0.772 & 0.040 & 0.255 \\
IFAN$_D$~\cite{Lee2021IFAN} & 10.5 & {28.66} & {0.868} & {0.025} & {0.172} & {23.46} & {0.743} & {0.049} & {0.240} & {25.99} & {0.804} & {0.037} & {0.207} \\
Restormer$_D$~\cite{Zamir2021Restormer} & 26.1 & {29.48}  & {0.895}  & \textcolor{blue}{0.023} & {0.134} & {23.97}  & \textcolor{blue}{0.773}  & \textcolor{blue}{0.047} & {0.175}  & \textcolor{blue}{26.66}  & \textcolor{blue}{0.833}  & \textcolor{blue}{0.035}  & {0.155} \\
Uformer$_D$~\cite{Uformer} & 20.6 & 28.23 & 0.860 & 0.026 & 0.199 & 23.10 & 0.728 & 0.051 & 0.285 & 25.65 & 0.795 & 0.039 & 0.243 \\
{GRL}$_D$-B~\cite{li2023grl} & 19.9 & \textcolor{blue}{29.83}  & \textcolor{blue}{0.903}  & \textcolor{red}{0.022} & \textcolor{red}{0.114} & \textcolor{blue}{24.39}  & \textcolor{red}{0.795}  & \textcolor{red}{0.045} & \textcolor{red}{0.150}  & \textcolor{blue}{27.04}  & \textcolor{blue}{0.847}  & \textcolor{red}{0.034}  & \textcolor{red}{0.133} \\
\textbf{TEAFormer}$_D$ \textbf{(Ours)} & \textbf{15.4} & \textbf{\textcolor{red}{30.16}}  & \textbf{\textcolor{red}{0.905}}  & \textbf{\textcolor{red}{0.022}} & \textbf{\textcolor{blue}{0.128}} & \textbf{\textcolor{red}{24.45}}  & \textbf{\textcolor{red}{0.795}}  & \textbf{\textcolor{red}{0.045}} & \textbf{\textcolor{blue}{0.170}}  & \textbf{\textcolor{red}{27.23}}  & \textbf{\textcolor{red}{0.849}}  & \textbf{\textcolor{red}{0.034}}  & \textcolor{blue}{\textbf{0.150}} \\
\bottomrule[0.15em]
\end{tabular}
}
\vspace{-3mm}
\caption{\textit{\textbf{Defocus deblurring}} comparisons on the DPDD~\cite{abuolaim2020defocus} testset (containing 37 indoor and 39 outdoor scenes). \textbf{S:} single-image defocus deblurring. \textbf{D:} dual-pixel defocus deblurring. The best two results are highlighted in \textcolor{red}{red} and \textcolor{blue}{blue}, respectively.}\label{tab:defocus_deblurring}
\vspace{-5mm}
\end{table*}

\noindent \textbf{Results.} We first compare TEAFormer with various classical image SR methods, including EDSR~\cite{lim2017enhanced}, RCAN~\cite{zhang2018rcan}, SwinIR~\cite{liang2021swinir}, EDT~\cite{ijcai2023p121}, HAT~\cite{chen2023hat}, SRFormer~\cite{zhou2023srformer}, GRL~\cite{li2023grl}, DAT~\cite{chen2023dual}, and IPG~\cite{tian2024image}. We compare them on synthetic benchmarks~\cite{Set5,Set14,BSD100,Urban100,Manga109} generated by bicubic downsampling. The FLOPs and latency reported were calculated with one NVIDIA A800 GPU on 768 resolution.

\begin{figure*}[!ht]
\centering
\newcommand{\h}{0.105}
\newcommand{\wa}{0.12}
\newcommand{\wb}{0.16}
\newcommand{\g}{-0.7mm}
\renewcommand{\tabcolsep}{1.8pt}
\renewcommand{\arraystretch}{1}
\resizebox{1.00\linewidth}{!} {
    \begin{tabular}{l}
        \normalsize
        \newcommand{\name}{images/dfb/}
        \renewcommand{\h}{0.086}
        \newcommand{\wq}{0.180}
        \begin{tabular}{cc}
        \normalsize
            \begin{adjustbox}{valign=t}
                \begin{tabular}{c}
                    \includegraphics[height=0.205\textwidth, width=0.374\textwidth]{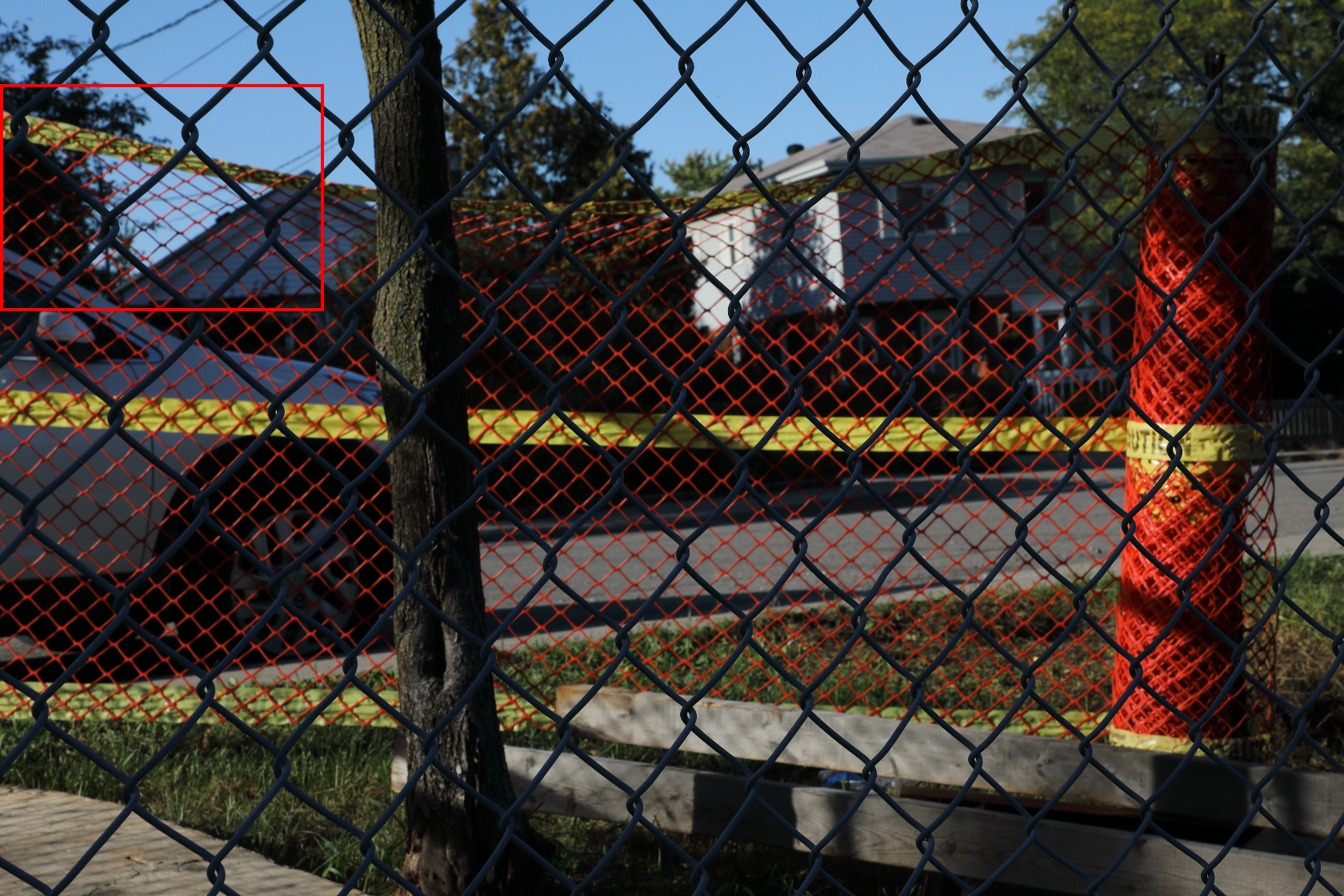}
                    \\
                    DPDD ($4\times$): img\_1651
                \end{tabular}
            \end{adjustbox}
            \begin{adjustbox}{valign=t}
                \begin{tabular}{ccccc}
                    \includegraphics[height=\h \textwidth, width=\wq \textwidth]{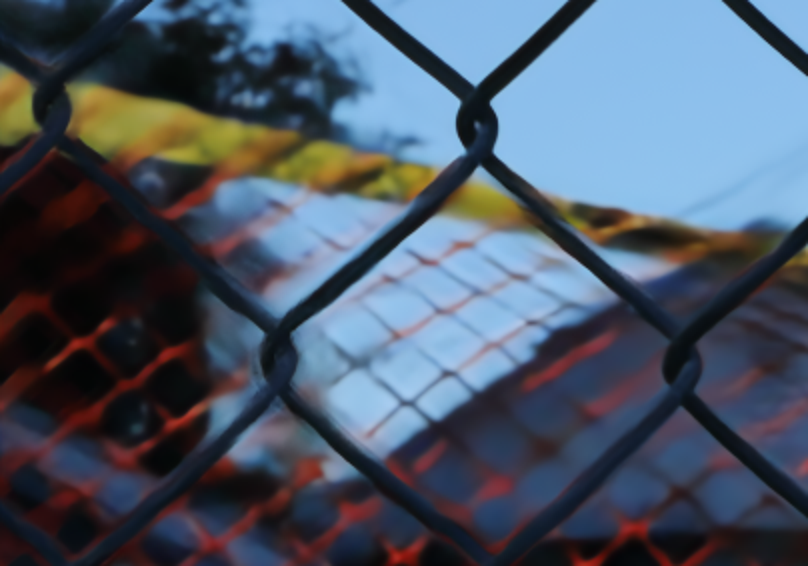} \hspace{\g} &
                    \includegraphics[height=\h \textwidth, width=\wq \textwidth]{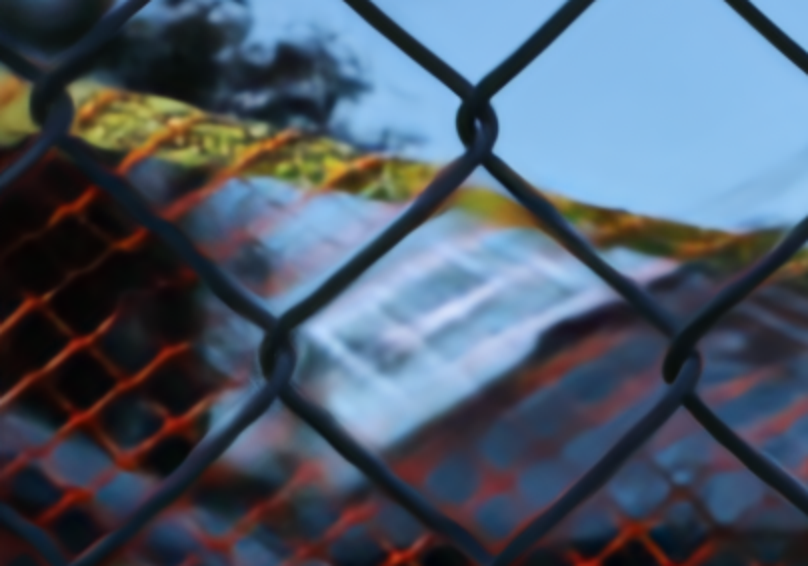} \hspace{\g} &
                    \includegraphics[height=\h \textwidth, width=\wq \textwidth]{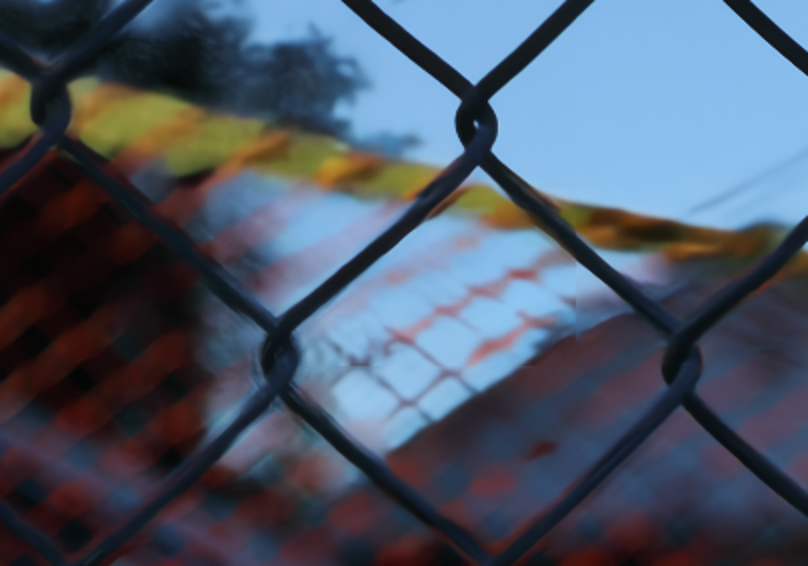} \hspace{\g} &
                    \includegraphics[height=\h \textwidth, width=\wq \textwidth]{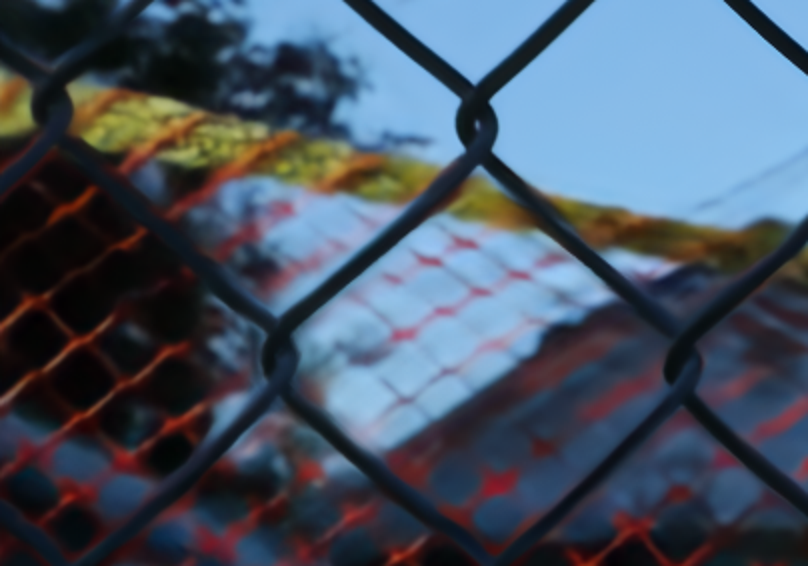} \hspace{\g} 
                    \\
                    Restormer$_S$~\cite{Zamir2021Restormer}\hspace{\g} &
                    SFNet$_S$~\cite{cui2023selective} \hspace{\g} &
                    GRL$_S$~\cite{li2023grl} \hspace{\g} &
                    \textbf{TEAFormer$_S$ (Ours)} \hspace{\g} 
                    \\
                    \vspace{-4.0mm}
                    \\
                    
                    \includegraphics[height=\h \textwidth, width=\wq \textwidth]{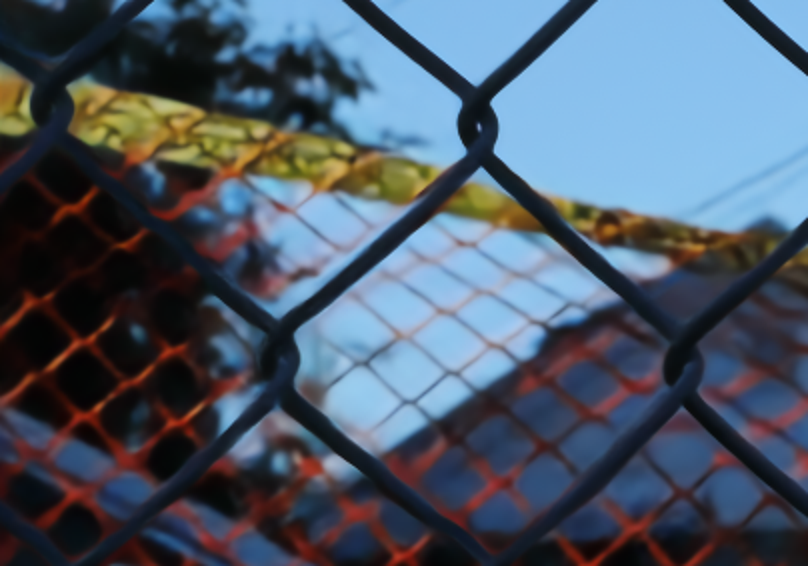} \hspace{\g} &
                    \includegraphics[height=\h \textwidth, width=\wq \textwidth]{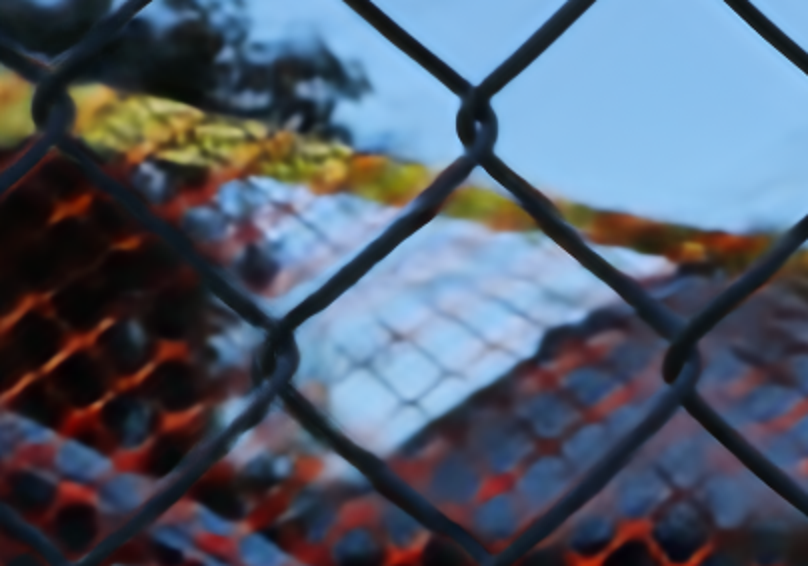} \hspace{\g} &
                    \includegraphics[height=\h \textwidth, width=\wq \textwidth]{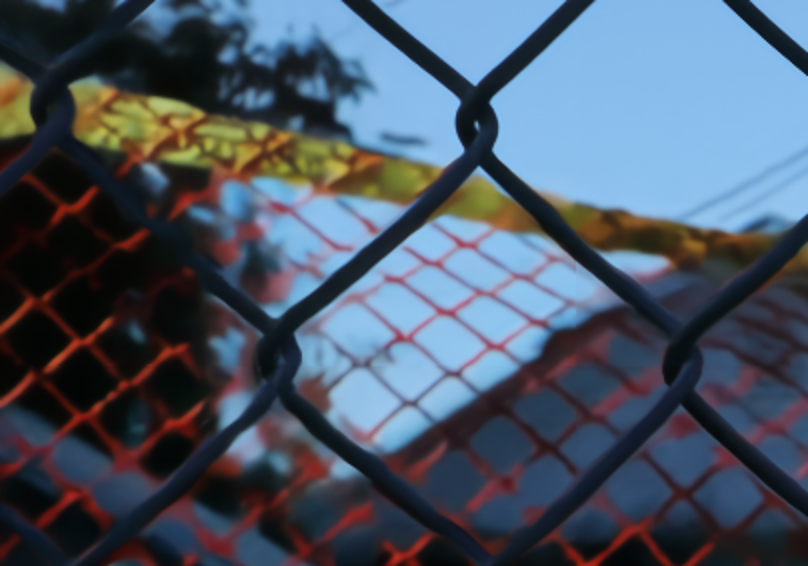} \hspace{\g} &
                    \includegraphics[height=\h \textwidth, width=\wq \textwidth]{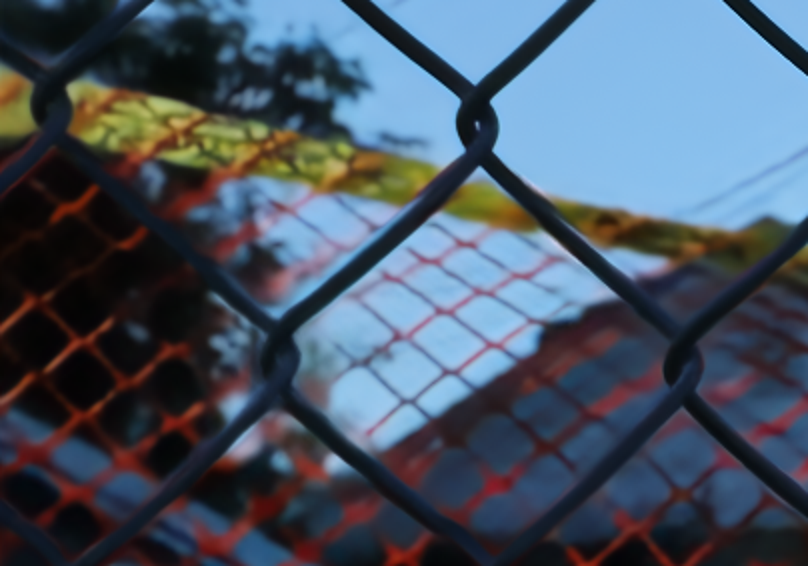} \hspace{\g} 
                    \\ 
                    Restormer$_D$~\cite{Zamir2021Restormer} &
                    SFNet$_D$~\cite{cui2023selective} &
                    GRL$_D$~\cite{li2023grl} &
                    \textbf{TEAFormer$_D$ (Ours)}  \\
                \end{tabular}
            \end{adjustbox}
        \end{tabular} \\
    \end{tabular}
}
\vspace{-4mm}
\caption{Qualitative comparison with recent SOTA methods on the \textit{\textbf{image defocus blur}} task. Please zoom in for better view.} \label{fig:dfb}
\vspace{-3mm}
\end{figure*}

The quantitative comparison of the methods for classical image SR is shown in Table~\ref{tab:super-resolution}. The exceptional performance of TEAFormer is distinctly evident as it achieves the highest results across nearly all five datasets for various scale factors. Incorporating the attention module with translation equivalence introduces a more suitable and productive inductive bias to the restoration model, thereby enhancing its performance. For the $\times4$ SR, our TEAFormer achieves a 28.67 dB PSNR score on the Urban100 dataset, which is 0.7 dB higher than HAT~\cite{chen2023hat} while both has 20M parameters. In comparison to the preceding image SR method IPG~\cite{tian2024image}, TEAFormer demonstrates a gain of 0.54 dB with comparable FLOPs and a $2\times$ faster inference. 

We further conduct TEAFormer-Light, which only comprises 829k parameters. In comparison to the CNN method HAN~\cite{niu2020single}, TEAFormer-Light has been demonstrated to achieve superior performance with only its 1.3\% parameters. In comparison to other lightweight transformer methods, TEAFormer also demonstrates an absolute performance advantage, as evidenced by the fact that TEAFormer-Light achieves a higher PSNR than SwinIR-L on the Urban100~\cite{Urban100} dataset by 0.57 dB.

\subsection{Image Defocus Deblurring}

\textbf{Training settings.} We employed the same training hyperparameters and evaluation as Restormer \cite{Zamir2021Restormer} in image deblurring tasks. Our TEAFormer were trained from scratch using progressive learning \cite{Zamir2021Restormer}. During the training phase, the optimizer adopted was AdamW~\cite{loshchilov2017decoupled} with $\text{L}_1$ loss.

\noindent \textbf{Results.} We compare our TEAFormer with single image defocus deblurring methods~\cite{abuolaim2020defocus,Son_2021_ICCV,Lee2021IFAN,Zamir2021Restormer,cui2023selective,li2023grl,IRNeXt} and dual-pixel image defocus deblurring methods~\cite{abuolaim2020defocus,abuolaim2021learning,Lee2021IFAN,Zamir2021Restormer,Uformer,li2023grl} on DPDD~\cite{abuolaim2020defocus} dataset in Table~\ref{tab:defocus_deblurring}. Our TEAFormer significantly outperforms state-of-the-art schemes for the single-image and dual-pixel defocus deblurring tasks on PSNR and SSIM metrics. In particular in the category of single-pixel indoor scene, TEAFormer yields 0.44 dB improvements over the GRL previous best transformer method~\cite{li2023grl}. Figure~\ref{fig:dfb} illustrates that our TEAFormer is more effective in removing spatially varying defocus blur than other restoration transformers.

\vspace{-5pt}
\subsection{Image Denoising}

\textbf{Training settings.} The image denoising and defocus deblurring tasks used the same hyper-parameters for training. 

\noindent \textbf{Gaussian Denoising Results.} We perform denoising experiments on synthetic benchmark datasets generated with additive white Gaussian noise (Set12~\cite{zhang2017beyond}, BSD68~\cite{bsd68}, Urban100~\cite{Urban100}, Kodak24~\cite{kodak24} and McMaster~\cite{zhang2011color}). Following DRUNet~\cite{zhang2021plug}, we learn a single model to handle various noise levels, including 15, 25, and 50. Our TEAFormer achieves state-of-the-art performance under both experimental settings on different datasets and noise levels. Specifically, for the challenging noise level 50 on color Urban100~\cite{Urban100} dataset, TEAFormer achieves 0.13 dB gain over the previous best restoration transformer GRL~\cite{li2023grl}, as shown in Table~\ref{tab:gaussian_denoising}. Similar performance gains can be observed for the Gaussian gray denoising. 

\begin{table*}[ht]
\centering
\setlength{\tabcolsep}{1.3pt}
\scalebox{0.73}{
\normalsize
\begin{tabular}{l | c | c c c | c c c | c c c | c c c || c c c | c c c | c c c }
\toprule[0.1em]
\multirow{3}{*}{\textbf{Method}} & \textbf{Params} & \multicolumn{12}{c||}{\textbf{Color}} & \multicolumn{9}{c}{\textbf{Grayscale}} \\ \cline{2-23}
&  & \multicolumn{3}{c|}{\textbf{CBSD68}~\cite{martin2001database}} & \multicolumn{3}{c|}{\textbf{Kodak24}~\cite{kodak24}} & \multicolumn{3}{c|}{\textbf{McMaster}~\cite{zhang2011color}} & \multicolumn{3}{c||}{\textbf{Urban100}~\cite{Urban100}}  & \multicolumn{3}{c|}{\textbf{Set12}~\cite{zhang2017beyond}} & \multicolumn{3}{c|}{\textbf{BSD68}~\cite{martin2001database}} & \multicolumn{3}{c}{\textbf{Urban100}~\cite{Urban100}} \\
& (M) & $\sigma$$=$$15$ & $\sigma$$=$$25$ & $\sigma$$=$$50$ & $\sigma$$=$$15$ & $\sigma$$=$$25$ & $\sigma$$=$$50$ & $\sigma$$=$$15$ & $\sigma$$=$$25$ & $\sigma$$=$$50$ & $\sigma$$=$$15$ & $\sigma$$=$$25$ & $\sigma$$=$$50$ & $\sigma$$=$$15$ & $\sigma$$=$$25$ & $\sigma$$=$$50$ & $\sigma$$=$$15$ & $\sigma$$=$$25$ & $\sigma$$=$$50$ & $\sigma$$=$$15$ & $\sigma$$=$$25$ & $\sigma$$=$$50$ \\ \midrule
DnCNN~\cite{zhang2017beyond}	&{0.56}	&33.90	&31.24	&27.95	&34.60	&32.14	&28.95	&33.45	&31.52	&28.62	&32.98	&30.81	&27.59	&32.86	&30.44	&27.18	&31.73	&29.23	&26.23	&32.64	&29.95	&26.26	\\
DRUNet~\cite{zhang2021plug}	& 32.6 &34.30	&31.69	&28.51	&35.31	&32.89	&29.86	&35.40	&33.14	&30.08	&34.81	&32.60	&29.61 &33.25	&30.94	&27.90	&31.91	&29.48	&26.59	&33.44	&31.11	&27.96	\\
\specialrule{0em}{1pt}{1pt}
\hdashline
\specialrule{0em}{1pt}{1pt}
EDT-B~\cite{ijcai2023p121}	&11.5	&34.39	&31.76	&28.56	&35.37	&32.94	&29.87	&{35.61}	&{33.34}	&30.25	&35.22	&{33.07}	&{30.16}				&-	&-	&-	&-	&-	&-	&-	&-	&-	\\
SwinIR*~\cite{liang2021swinir}	&11.7	&{34.39}	&31.78	&28.55	&35.33	&32.88	&29.77	&{35.59}	&33.20	&30.19	&35.11	&32.89	&29.77	&33.34	&31.00	&27.88	&\textcolor{blue}{31.96}	&29.49	&26.58	&33.61	&31.20	&27.86	\\
Restormer~\cite{Zamir2021Restormer}	&26.1	&34.39	&{31.78}	&\textcolor{blue}{28.59}	&{35.44}	&{33.02}	&{30.00}	&{35.55}	&{33.31}	&{30.29}	&35.06	&32.91	&30.02				&\textcolor{blue}{33.35}	&{31.04}	&\textcolor{blue}{28.01}	&31.95	&\textcolor{blue}{29.51}	&\textcolor{blue}{26.62}	&33.67	&31.39	&{28.33}	\\
ART*~\cite{zhang2023accurate}	&16.2	&\textcolor{blue}{34.40}	&\textcolor{blue}{31.79}	&\textcolor{blue}{28.59}	&\textcolor{blue}{35.47}	&\textcolor{blue}{33.04}	&{30.00}	&{35.61}	&{33.33}	&\textcolor{blue}{30.27}	&35.12	&32.95	&29.99	&\textcolor{blue}{33.35}	&\textcolor{blue}{31.05}	&\textcolor{blue}{28.01}	&{31.95}	&{29.50}	&{26.59}	&{33.71}	&{31.47}	&{28.40}	\\
GRL-B*~\cite{li2023grl}	&19.8	&\textcolor{blue}{34.40}	&{31.78}	&{28.58}	&{35.42}	&{33.03}	&\textcolor{blue}{30.02}	&\textcolor{blue}{35.63}	&\textcolor{blue}{33.36}	&{30.26}	&\textcolor{blue}{35.22}	&\textcolor{blue}{33.01}	&\textcolor{blue}{30.24}				&{33.34}	&{31.02}	&{27.99}	&{31.95}	&{29.50}	&{26.60}	&\textcolor{blue}{33.73}	&\textcolor{blue}{31.54}	&\textcolor{blue}{28.49}	\\	
\textbf{TEAFormer (Ours)}	&\textbf{15.4}	&\textcolor{red}{\textbf{34.40}}	&\textcolor{red}{\textbf{31.80}}	&\textcolor{red}{\textbf{28.62}}	&\textcolor{red}{\textbf{35.50}}	&\textcolor{red}{\textbf{33.08}}	&\textcolor{red}{\textbf{30.05}}	&\textcolor{red}{\textbf{35.68}}	&\textcolor{red}{\textbf{33.42}}	&\textcolor{red}{\textbf{30.38}}	&\textcolor{red}{\textbf{35.25}}	&\textcolor{red}{\textbf{33.18}}	&\textcolor{red}{\textbf{30.37}}	&\textcolor{red}{\textbf{33.37}}	&\textcolor{red}{\textbf{31.06}}	&\textcolor{red}{\textbf{28.04}}	&\textcolor{red}{\textbf{31.96}}	&\textcolor{red}{\textbf{29.52}}	&\textcolor{red}{\textbf{26.63}}	&\textcolor{red}{\textbf{33.89}}	&\textcolor{red}{\textbf{31.65}}	&\textcolor{red}{\textbf{28.61}}	\\
\bottomrule[0.1em]
\end{tabular}}
\vspace{-3mm}
\caption{\textit{\textbf{Color and grayscale image Gaussian denoising}} results in terms of PSNR $\uparrow$. Model complexity and prediction accuracy are shown for better comparison. Dash lines separate the results of CNNs and Transformers. The best two results are highlighted in \textcolor{red}{red} and \textcolor{blue}{blue}, respectively. Methods with ``*" are replicated with standard training setting, which learns a single model to handle various noise levels.}\label{tab:gaussian_denoising}
\vspace{-7mm}
\end{table*}


\subsection{All-in-one Image Restoration}

\begin{table*}[ht]
\setlength\tabcolsep{5pt}
\centering
\scalebox{0.75}{
\begin{tabular}{l|c|c|c|c|c|c|c}
\toprule[0.15em]
\multirow{3}{*}{\textbf{Method}} & \multirow{2}{*}{\textbf{Params}} & \textbf{Dehazing} & \textbf{Deraining} & \textbf{Denoising} & \textbf{Deblurring} & \textbf{Low-Light}  & \multirow{2}{*}{\textbf{Average}} \\ 
~ &  & on SOTS~\citep{RESIDE} & on Rain100L~\cite{rain100L} & on BSD68~\citep{bsd68} & on GoPro~\cite{gopro2017} & on LOL~\citep{Chen2018Retinex} & ~ \\ 
\specialrule{0em}{1pt}{1pt}
\cline{2-8}
\specialrule{0em}{1pt}{1pt}
~ & (M) & PSNR$\uparrow$ / SSIM$\uparrow$ & PSNR$\uparrow$ / SSIM$\uparrow$ & PSNR$\uparrow$ / SSIM$\uparrow$ & PSNR$\uparrow$ / SSIM$\uparrow$ & PSNR$\uparrow$ / SSIM$\uparrow$ & PSNR$\uparrow$ / SSIM$\uparrow$ \\
\midrule[0.15em]
AirNet~\cite{airnet} & 8.90& 21.04 / 0.884 &32.98 / 0.951 &30.91 / 0.882 &24.35 / 0.781 &18.18 / 0.735 &25.49 / 0.846\\
IDR~\cite{zhang2023ingredient} & 15.3& {25.24} / {0.943}&35.63 / 0.965 & \textcolor{red}{31.60} / \textcolor{blue}{0.887} &{27.87} / {0.846} &21.34 / {0.826} &{28.34} / {0.893} \\
PromptIR~\citep{promptir} & 35.6& {25.20} / {0.931} & {35.94} / {0.964} & {31.17} / {0.882} & {27.32} / {0.842} & {20.94} / {0.799} & {28.11} / {0.883} \\ 
InstructIR~\cite{conde2024instructir} & 35.1& \textcolor{blue}{27.10} / \textcolor{blue}{0.956} & \textcolor{blue}{36.84} / \textcolor{blue}{0.973} & 31.40 / \textcolor{blue}{0.887} & \textcolor{blue}{29.40} / \textcolor{blue}{0.886} & \textcolor{blue}{23.00} / \textcolor{blue}{0.836} & \textcolor{blue}{29.55} / \textcolor{blue}{0.907} \\
\specialrule{0em}{1pt}{1pt}
\hdashline
\specialrule{0em}{1pt}{1pt}
SwinIR~\cite{liang2021swinir} & 11.8& 21.50 / 0.891& 30.78 / 0.923  &30.59 / 0.868& 24.52 / 0.773 &17.81 / 0.723& 25.04 / 0.835 \\
NAFNet~\cite{nafnet} & 67.9& 25.23 / 0.939 &35.56 / 0.967 &31.02 / 0.883 &26.53 / 0.808 &20.49 / 0.809& 27.76 / 0.881 \\
Restormer~\cite{Zamir2021Restormer} & 26.1& 24.09 / 0.927 &34.81 / 0.962 &{31.49} / 0.884 &27.22 / 0.829 &20.41 / 0.806 &27.60 / 0.881 \\
\textbf{TEAFormer~(Ours)} & \textbf{15.4} & \textcolor{red}{\textbf{31.57}} / \textcolor{red}{\textbf{0.980}} & \textcolor{red}{\textbf{39.44}} / \textcolor{red}{\textbf{0.986}} & \textcolor{blue}{\textbf{31.54}} / \textcolor{red}{\textbf{0.890}} & \textcolor{red}{\textbf{30.53}} / \textcolor{red}{\textbf{0.908}} & \textcolor{red}{\textbf{23.06}} / \textcolor{red}{\textbf{0.856}} & \textcolor{red}{\textbf{31.22}} / \textcolor{red}{\textbf{0.924}} \\ 
\bottomrule[0.15em]
\end{tabular}
}
\vspace{-3mm}
\caption{\textbf{\textit{5D All-in-one image restoration results.}} Dash lines separate the results of methods designed specifically for all-in-one image restoration task and methods that only focus on architecture enhancement. The best two results are highlighted in \textcolor{red}{red} and \textcolor{blue}{blue}, respectively. TEAFormer sets the new state-of-the-art performance in each task.}\label{tab:all_in_one_5d}
\end{table*}


\noindent \textbf{Training settings.} Following DCPT~\cite{hu2025universal}, a 5D all-in-one image restoration and enhancement task was conducted with the objective of training the network to eliminate five distinct types of degradation: haze, rain, Gaussian noise, motion blur, and low light. The models were trained from scratch on datasets with $128\times128$ resolution without progressive learning. During training, the optimizer was set as AdamW, the learning rate was set as $3\times10^{-4}$ while using the cosine scheduler. We used $L_1$ loss as a criterion.

\noindent \textbf{Results.} As illustrated in Table~\ref{tab:all_in_one_5d}, TEAFormer surpasses the most recently proposed methods specifically designed for comprehensive all-in-one restoration tasks (IDR~\cite{zhang2023ingredient}, PromptIR~\cite{promptir}, and InstructIR~\cite{conde2024instructir}). The average PSNR of TEAFormer in all five tasks reached 31.22 dB, thereby exceeding the current state-of-the-art 5D all-in-one restoration method, InstructIR~\cite{conde2024instructir}, by 1.67 dB. TEAFormer consistently outperforms prior PSNR in every indicator of each task, signifying its exceptional proficiency in image restoration under complex multi-task conditions.


\begin{table*}[!ht]
\parbox{.46\linewidth}{
\centering
\vspace{-3mm}
\setlength{\tabcolsep}{1.8pt}
\scalebox{0.72}{
\normalsize
\begin{tabular}{ccc}
\includegraphics[width=0.42\linewidth]{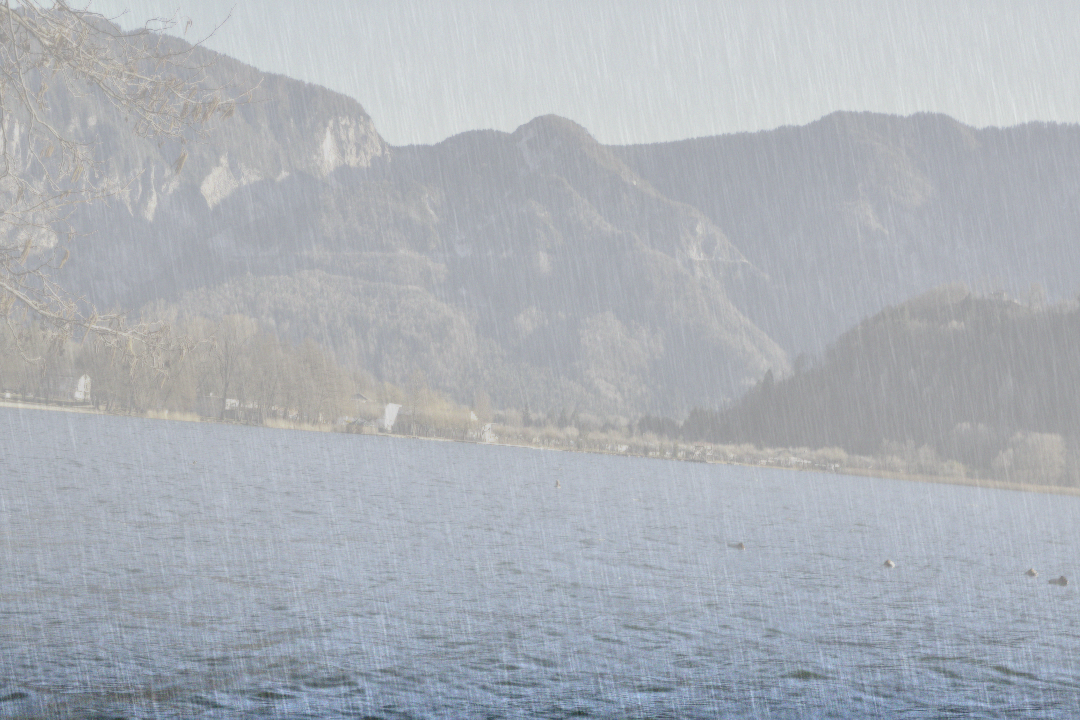} & \includegraphics[width=0.42\linewidth]{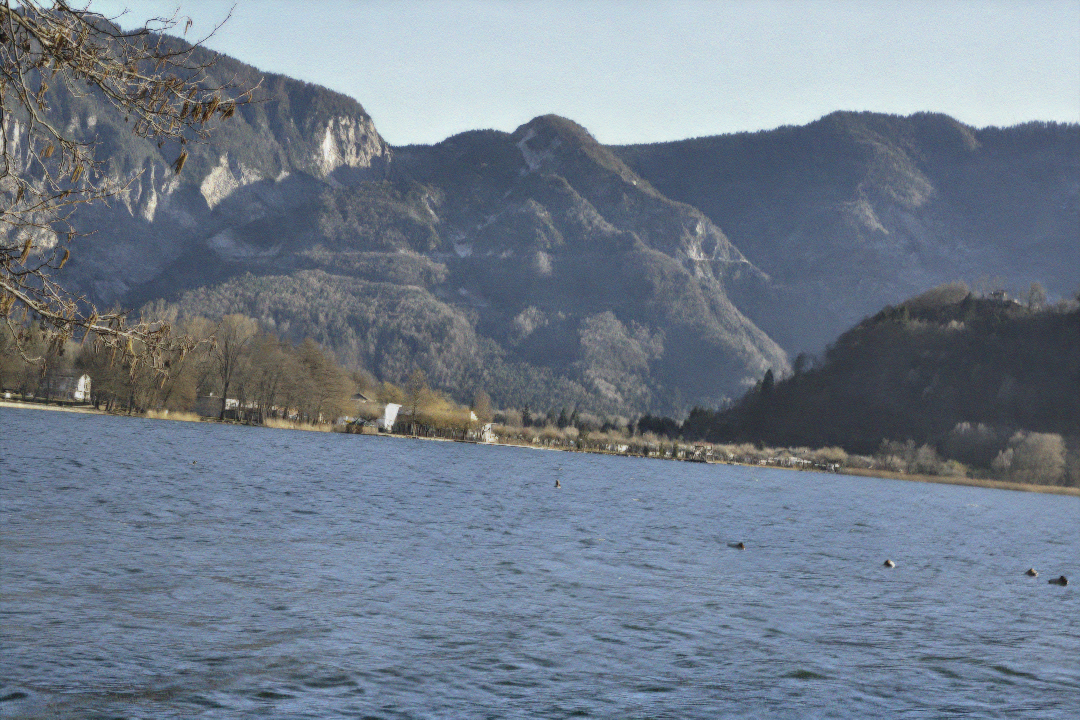} & \includegraphics[width=0.42\linewidth]{images/cdd/onerestore.png} \\
LQ & InsturctIR~\cite{conde2024instructir} & OneRestore~\cite{guo2024onerestore} \\
\includegraphics[width=0.42\linewidth]{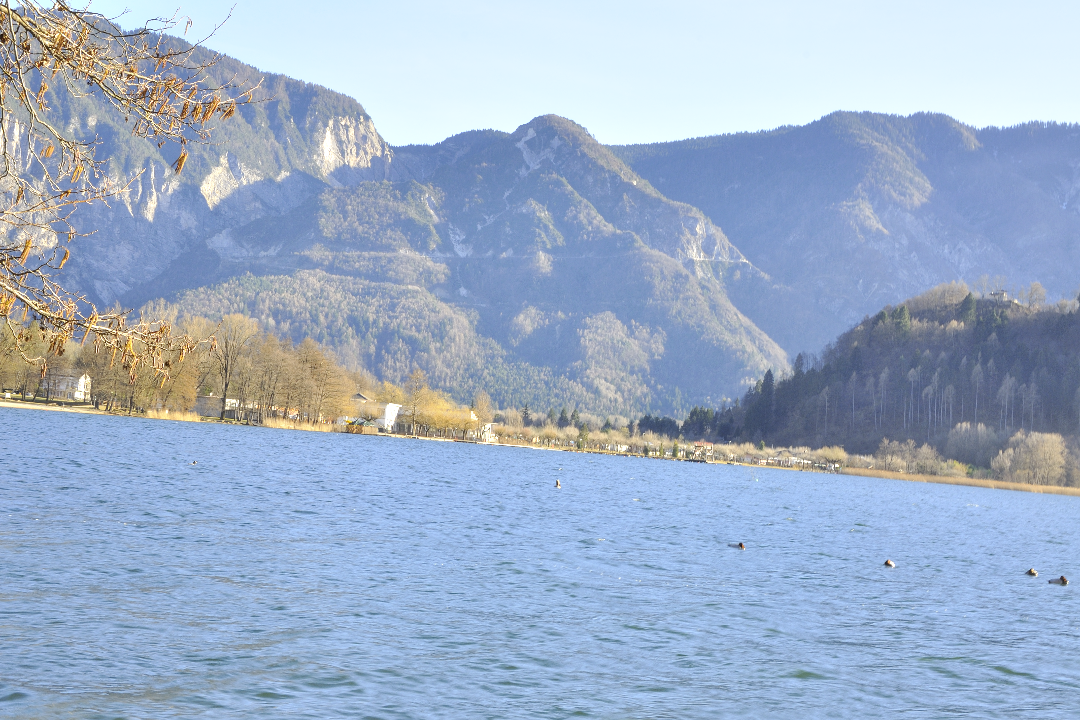} & \includegraphics[width=0.42\linewidth]{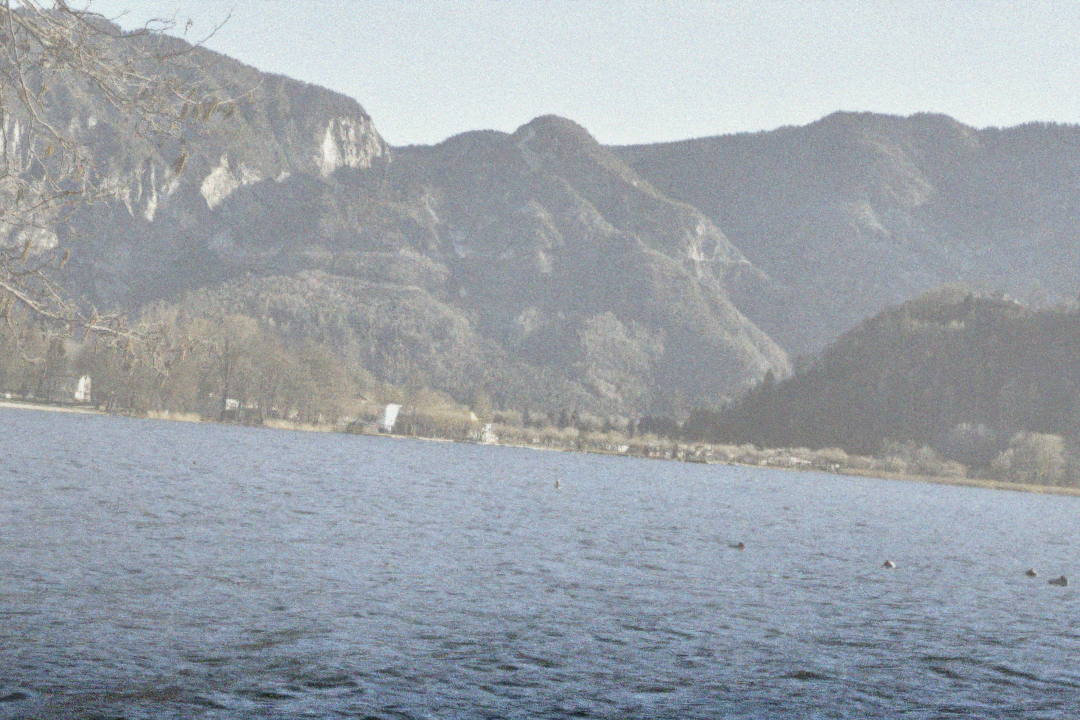} & \includegraphics[width=0.42\linewidth]{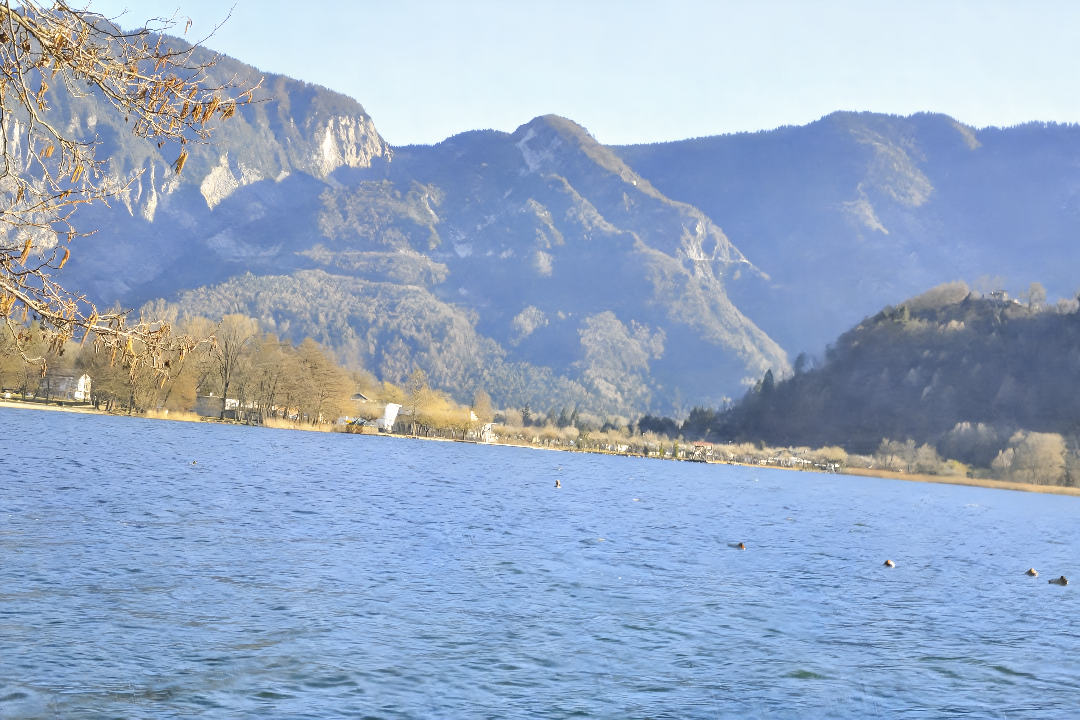} \\
HQ & DACLIP~\cite{luo2023controlling} & \textbf{TEAFormer (Ours)} \\
\end{tabular}
}
\vspace{-3mm}
\captionof{figure}{\small \textbf{\textit{Visual comparison on low-light+haze+rain mixed degradation.}} TEAFormer can restore the correct lightness.}\label{fig:cdd}
\vspace{-4mm}
}
\hfill
\parbox{.54\linewidth}{
\centering
\vspace{-3mm}
\setlength{\tabcolsep}{1.8pt}
\scalebox{0.85}{
\begin{tabular}{l|c|c|c}
\toprule[0.15em]
\multirow{2}{*}{\textbf{Method}} & \textbf{Params} & \textbf{low-light+haze+rain} & \textbf{low-light+haze+snow} \\ 
\specialrule{0em}{1pt}{1pt}
\cline{2-4}
\specialrule{0em}{1pt}{1pt}
~ & (M) & PSNR$\uparrow$ / SSIM$\uparrow$ & PSNR$\uparrow$ / SSIM$\uparrow$ \\
\midrule[0.15em]
AirNet~\cite{airnet} & 8.90 & 21.80 / 0.708 & 22.23 / 0.725 \\
TransWeather~\cite{valanarasu2022transweather} & 31.0 & 21.55 / 0.678 & 21.01 / 0.655 \\
PromptIR~\cite{promptir} & 35.6 & 23.74 / 0.752 & 23.33 / 0.747 \\
DACLIP~\cite{luo2023controlling} & 174 & 22.96 / 0.712 & 23.32 / 0.751 \\
DiffUIR-L~\cite{zheng2024selective} & 36.3 & 25.46 / 0.779 & 25.89 / 0.788 \\
OneRestore~\cite{guo2024onerestore} & 5.98 & 25.18 / 0.795 & 25.28 / 0.797 \\
InstructIR~\cite{conde2024instructir}& 35.1 & 24.84 / 0.777 & 24.32 / 0.760 \\
\textbf{TEAFormer (Ours)} & \textbf{15.4} & \textbf{25.88} / \textbf{0.796} & \textbf{26.05} / \textbf{0.800} \\ 
\bottomrule[0.15em]
\end{tabular}
}
\vspace{-3mm}
\caption{\small \textbf{\textit{Mixed degraded image restoration results}} on CDD~\citep{guo2024onerestore}.}\label{tab:mixed}
\vspace{-4mm}
}
\end{table*}

\noindent \textbf{Mixed degradation.} As shown in Table~\ref{tab:mixed}, TEAFormer surpasses the most recently proposed methods under mixed degradation scenarios, specifically designed for universal restoration tasks~\cite{luo2023controlling,zheng2024selective}). Within scenarios characterized by a combination of three degradation (low-light, haze, and rain), TEAFormer demonstrates superior performance over the current state-of-the-art universal restoration method, DiffUIR~\cite{zheng2024selective}, by a margin of 0.42 dB. 


\vspace{-5pt}
\subsection{Discussion}\label{sec:ablation}

\noindent \textbf{Hyperparameters analysis.} The performance under various hyperparameters is reported in Table~\ref{tab:ablation_w} ($w$), Table~\ref{tab:ablation_s} ($s$), Table~\ref{tab:ablation_k} ($k$) and Table~\ref{tab:ablation_n_d} ($N_d$). Our settings ($w=15,s=4,k=3,N_d=16$) optimally balanced performance and computation cost. Note that changing some of the hyperparameters does not affect the computational complexity as discussed in Section~\ref{sec:stacking}. For $s$ in Table~\ref{tab:ablation_s}, parameters and FLOPs are omitted because $s$ does not affect them.

\begin{table*}[!ht]
\begin{minipage}[t]{0.19\textwidth}
\centering
\normalsize
\setlength{\tabcolsep}{3pt}
\scalebox{0.75}{
\begin{tabular}{c|c|c}
\toprule[0.15em]
\multirow{2}{*}{$w$} & FLOPs & {Urban100~\cite{Urban100}} \\
\specialrule{0em}{1pt}{1pt}
\cline{2-3}
\specialrule{0em}{1pt}{1pt}
& (T) & PSNR $\uparrow$ / SSIM $\uparrow$ \\
\midrule[0.15em]
7 & 0.827 & 28.35 / 0.8441 \\
\midrule
15 & 1.035 & 28.67 / 0.8489 \\
\midrule
31 & 1.866 & 28.73 / 0.8493 \\
\bottomrule[0.15em]
\end{tabular}
}
\vspace{-3mm}
\caption{Ablations of sliding window size $w$.}\label{tab:ablation_w}
\vspace{-7mm}
\end{minipage}
\begin{minipage}[t]{0.005\textwidth}
\end{minipage}
\begin{minipage}[t]{0.13\textwidth}
\centering
\normalsize
\setlength{\tabcolsep}{3pt}
\scalebox{0.75}{
\begin{tabular}{c|c}
\toprule[0.15em]
\multirow{2}{*}{$s$} & {Urban100~\cite{Urban100}} \\
\specialrule{0em}{1pt}{1pt}
\cline{2-2}
\specialrule{0em}{1pt}{1pt}
& PSNR $\uparrow$ / SSIM $\uparrow$ \\
\midrule[0.15em]
1 & 28.43 / 0.8431 \\
\midrule
2 & 28.51 / 0.8453 \\
\midrule
4 & 28.67 / 0.8489 \\
\bottomrule[0.15em]
\end{tabular}
}
\vspace{-3mm}
\caption{Ablations of sliding stride $s$.}\label{tab:ablation_s}
\vspace{-7mm}
\end{minipage}
\begin{minipage}[t]{0.005\textwidth}
\end{minipage}
\begin{minipage}[t]{0.2\textwidth}
\centering
\normalsize
\setlength{\tabcolsep}{3pt}
\scalebox{0.75}{
\begin{tabular}{c|c|c}
\toprule[0.15em]
\multirow{2}{*}{$k$} & FLOPs & Urban100~\cite{Urban100} \\
\specialrule{0em}{1pt}{1pt}
\cline{2-3}
\specialrule{0em}{1pt}{1pt}
 & (T) & PSNR $\uparrow$ / SSIM $\uparrow$ \\
\midrule[0.15em]
None & 0.966 & 28.49 / 0.8470 \\
\midrule
1 & 1.028 & 28.53 / 0.8471 \\
\midrule
3 & 1.035 & 28.67 / 0.8489 \\
\bottomrule[0.15em]
\end{tabular}
}
\vspace{-3mm}
\caption{Ablations of kernel size $k$ for adaptive indices. ``None" stands for no adaptive indices used.}\label{tab:ablation_k}
\vspace{-7mm}
\end{minipage}
\begin{minipage}[t]{0.005\textwidth}
\end{minipage}
\begin{minipage}[t]{0.2\textwidth}
\centering
\normalsize
\setlength{\tabcolsep}{3pt}
\scalebox{0.75}{
\begin{tabular}{c|c|c}
\toprule[0.15em]
\multirow{2}{*}{$N_d$} & FLOPs & Urban100~\cite{Urban100} \\
\specialrule{0em}{1pt}{1pt}
\cline{2-3}
\specialrule{0em}{1pt}{1pt}
& (T) & PSNR $\uparrow$ / SSIM $\uparrow$ \\
\midrule[0.15em]
None & 0.757 & 28.47 / 0.8457 \\
\midrule
16 & 1.035 & 28.67 / 0.8489 \\
\midrule
$N$ & 18.36 & 28.80 / 0.8503 \\
\bottomrule[0.15em]
\end{tabular}
}
\vspace{-3mm}
\caption{Ablations of $N_d$ in DSA. ``None" stands for no DSA used.}\label{tab:ablation_n_d}
\vspace{-7mm}
\end{minipage}
\begin{minipage}[t]{0.005\textwidth}
\end{minipage}
\begin{minipage}[t]{0.22\textwidth}
\centering
\normalsize
\setlength{\tabcolsep}{3pt}
\scalebox{0.75}{
\begin{tabular}{c|c|c}
\toprule[0.15em]
\multirow{2}{*}{Down} & S-conv & {Urban100~\cite{Urban100}} \\
\specialrule{0em}{1pt}{1pt}
\cline{2-3}
\specialrule{0em}{1pt}{1pt}
& (\%) $\uparrow$ & PSNR $\uparrow$ / SSIM $\uparrow$ \\
\midrule[0.15em]
MaxPool & 87.3 & 28.61 / 0.8477 \\
\midrule
LPD & 91.7 & 28.64 / 0.8493 \\
\midrule
AvgPool & 91.2 & 28.67 / 0.8489 \\
\bottomrule[0.15em]
\end{tabular}
}
\vspace{-3mm}
\caption{Ablations of downsamplers in DSA. The higher the S-Conv, the more strictly the model meets TE.}\label{tab:ablation_down}
\vspace{-7mm}
\end{minipage}
\end{table*}

\noindent \textbf{Compare different downsampler in DSA.} The incorporation of DSA into our TEA substantially enhances the performance of the model without adding undue complexity. However, the average pooling employed in DSA does not strictly adhere to the property of TE. Using a comparative ablation of several downsampling modules in Table~\ref{tab:ablation_down}, we experimentally demonstrate that the use of average pooling does not compromise the model's performance or its TE.

\begin{figure}
\centering
\includegraphics[width=0.95\linewidth]{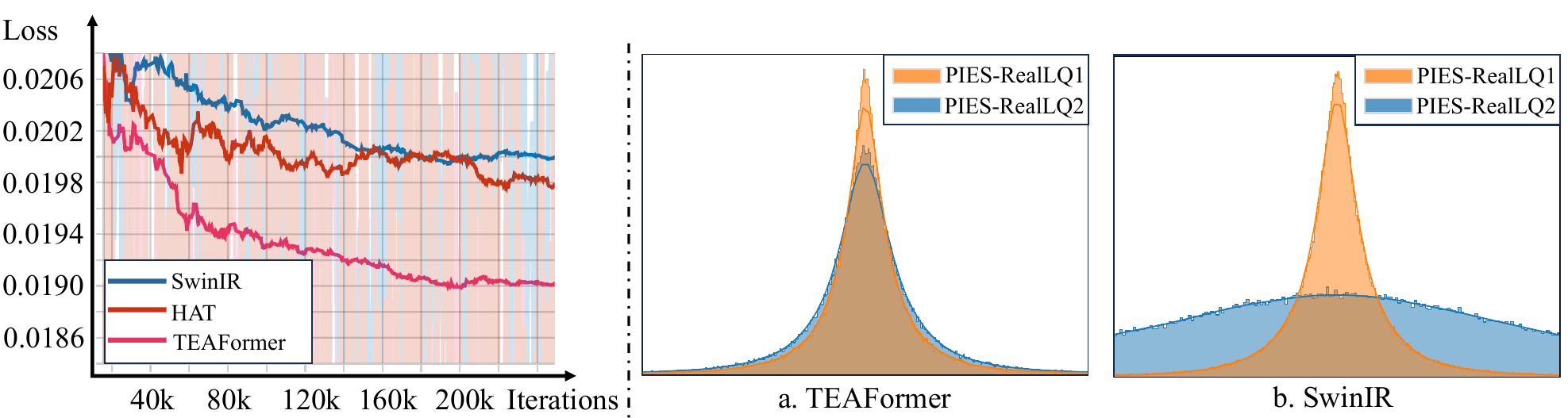}
\vspace{-3mm}
\caption{\footnotesize{Comparison on convergence (left) and generalization (right).}}
\label{fig:discussion}
\vspace{-5mm}
\end{figure}

\vspace{-3mm}
\noindent \textbf{Does TEAFormer converge faster and generalize better in practice?} Table~\ref{tab:te_ops} and Table~\ref{tab:ablation_tea} have illustrated that TEAFormer converges faster and generalizes better than other networks theoretically due to the incorporation of translation equivariance. Figure~\ref{fig:discussion} (left )further demonstrates that TEAFormer reaches a lower loss faster than other restoration transformers~\cite{liang2021swinir,chen2023hat} in practice. Following MaskDenoise~\cite{chen2023masked}, we compare the statistics of the last layer feature in SwinIR~\cite{liang2021swinir} and TEAFormer under two different real-world degradations on the PIES~\cite{liu2023evaluating} dataset (out of the training dataset) in Figure~\ref{fig:discussion} (right). The results show that TEAFormer can produce a more consistent output when processing LQ images with different degradations. Therefore, TEAFormer generalizes better in practice.

\vspace{-7pt}
\section{Conclusion}
\vspace{-1mm}

In this paper, we start by examining the translation equivariance, which serves as an inherent inductive bias in image restoration. We then define the translation equivariance and offer two strategies for incorporating it into restoration networks: (1) slide indexing and (2) component stacking.  To address the fixed receptive field provided by these strategies, we proposed TEA and TEAFormer via adaptive key-value indexing. The efficacy of TEAFormer has been demonstrated in various image restoration tasks, with ablation studies confirming that this efficacy arises from the incorporation of TEA. In the future, we intend to explore more equivariances into restoration networks.

\section*{Acknowledgement}
This work was supported financially the Natural Science Foundation of China (82371112, 623B2001), the Science Foundation of Peking University Cancer Hospital (JC202505), Natural Science Foundation of Beijing Municipality (Z210008) and the Clinical Medicine Plus X - Young Scholars Project of Peking University, the Fundamental Research Funds for the Central Universities (PKU2025PKULCXQ008).
{
    \small
    \bibliographystyle{ieeenat_fullname}
    \bibliography{main}
}
\clearpage
\setcounter{page}{1}
\maketitlesupplementary

\section*{A. Translation equivariance v.s. translation invariance}

In this section, we briefly explore the difference between translation equivariance v.s. translation invariance.

\noindent \textbf{Definition A.1. Translation Equivariance.} We call function $\Phi(\cdot)$ is translation equivariant to translation operation $\mathcal{T}(\cdot)$, if $\Phi(\mathcal{T}(x)) = \mathcal{T}(\Phi(x))$, where $x$ is the input signal. 

\noindent \textbf{Definition A.2. Translation Invariance.} We call function $\Phi(\cdot)$ is translation invariant to translation operation $\mathcal{T}(\cdot)$, if $\Phi(\mathcal{T}(x)) = \Phi(x)$, where $x$ is the input signal. 

According to the aforementioned definitions, it can be inferred that a translation-equivariant operator ensures that when the input undergoes a translation, the output is translated accordingly. Equivariance maintains the spatial correspondence between the input and output. Conversely, a translation-invariant operator guarantees that the output remains unchanged when the input is translated, thereby ensuring stability against translation. Translation invariance highlights robustness to input variations. 

In the context of image restoration tasks, where precise restoration is required at the pixel level, the model should maintain a degree of translation equivariance rather than translation invariance to enhance the fidelity of the restored image as we discussed in ~\cref{sec:intro}.

\section*{B. Proofs}

In this section, we provide proofs of Theorem 3.2 and Theorem 3.3.

\noindent \textbf{Proof of Theorem 3.2.}

\noindent \textit{Proof.} Given a function $\Phi(x)_i$ is transformed from $x_j = [i-b,i+b]$, where $b$ is the sliding boundary, $\Phi(x)_i$ can be rewritten as follows.

\begin{equation}
\Phi(x)_i = \Phi(x_{j=[i - b,i + b]}),
\end{equation}

\noindent where $j$ is the index of the input signal $x$.

Given the translation operator $\mathcal{T}(\cdot)$, which satisfies $\mathcal{T}(x_j) = x_{j+\delta}$, where $\delta$ is a constant scalar.

\begin{equation}
\begin{aligned}
\mathcal{T}(\Phi(x)_i) &= \Phi(x)_{i + \delta} \\
&= \Phi(x_{j=[i + \delta - b,i + \delta + b]}) \\
&= \Phi(\mathcal{T}(x)_{j=[i - b,i + b]}) \\
&= \Phi(\mathcal{T}(x))_i,
\end{aligned}
\end{equation}

\noindent which completes the proof.

\noindent \textbf{Proof of Theorem 3.3.}

\noindent \textit{Proof.} Given functions $\Phi_1(\cdot)$ and $\Phi_2(\cdot)$ are translation equivariant to translation operation $\mathcal{T}(\cdot)$, according to Definition 3.1, we get:

\begin{align}
\Phi_1(\mathcal{T}(x)) = \mathcal{T}(\Phi_1(x)), \\
\Phi_2(\mathcal{T}(x)) = \mathcal{T}(\Phi_2(x)).
\end{align}

Since the translation operation $\mathcal{T}(\cdot)$ is a linear operator, we can sum the above equations as follows.

\begin{equation}
\begin{aligned}
\Phi_1(\mathcal{T}(x)) + \Phi_2(\mathcal{T}(x)) &= \mathcal{T}(\Phi_1(x)) + \mathcal{T}(\Phi_2(x)) \\
&= \mathcal{T}(\Phi_1(x) + \Phi_2(x)).
\end{aligned}
\end{equation}

It can be seen that the function $\Phi_1(\cdot) + \Phi_2(\cdot)$ is translation equivariant to translation operation $\mathcal{T}(\cdot)$.

When stacking the functions $\Phi_1(\cdot)$ and $\Phi_2(\cdot)$ in parallel,

\begin{equation}
\begin{aligned}
\Phi_2(\Phi_1(\mathcal{T}(x))) &= \Phi_2(\mathcal{T}(\Phi_1(x)) \\
&= \mathcal{T}(\Phi_2(\Phi_1(x))).
\end{aligned}
\end{equation}

The function $\Phi_2(\Phi_1(\cdot))$ is also translation equivariant to translation operation $\mathcal{T}(\cdot)$, which completes the proof.

\end{document}